\documentclass[letterpaper]{article} 
\usepackage{aaai23}  
\usepackage{times}  
\usepackage{helvet}  
\usepackage{courier}  
\usepackage[hyphens]{url}  
\usepackage{graphicx} 
\usepackage{natbib}  
\usepackage{caption} 
\usepackage{algorithm}
\usepackage{algorithmic}

\usepackage{newfloat}
\usepackage{listings}
\DeclareCaptionStyle{ruled}{labelfont=normalfont,labelsep=colon,strut=off} 
\floatstyle{ruled}
\newfloat{listing}{tb}{lst}{}
\floatname{listing}{Listing}
\usepackage{caption}
\usepackage{subcaption}
\usepackage{booktabs}
\usepackage{multirow}

\title{Demystifying Randomly Initialized Networks for Evaluating Generative Models}
\author {
    Junghyuk Lee, 
    Jun-Hyuk Kim, 
    Jong-Seok Lee
}
\affiliations {
    Yonsei University, South Korea\\
    \{junghyuklee, junhyuk.kim, jong-seok.lee\}@yonsei.ac.kr
}

\usepackage{bibentry}

\begin{document}

\maketitle

\newcommand{\Frechet}[0]{Fr\'echet~}
\newcommand{\pokemon}[1][~]{Pok\'emon#1}
\newcommand{\specialcell}[2][c]{%
  \begin{tabular}[#1]{@{}c@{}}#2\end{tabular}}
\newcommand{\specialcellr}[2][c]{%
  \begin{tabular}[#1]{@{}#1@{}}#2\end{tabular}}
\newcommand{\alignc}[1]{\multicolumn{1}{c}{#1}}

\begin{abstract}
Evaluation of generative models is mostly based on the comparison between the estimated distribution and the ground truth distribution in a certain feature space. To embed samples into informative features, previous works often use convolutional neural networks optimized for classification, which is criticized by recent studies. Therefore, various feature spaces have been explored to discover alternatives. Among them, a surprising approach is to use a randomly initialized neural network for feature embedding. However, the fundamental basis to employ the random features has not been sufficiently justified. In this paper, we rigorously investigate the feature space of models with random weights in comparison to that of trained models. Furthermore, we provide an empirical evidence to choose networks for random features to obtain consistent and reliable results. Our results indicate that the features from random networks can evaluate generative models well similarly to those from trained networks, and furthermore, the two types of features can be used together in a complementary way.
\end{abstract}

\section{Introduction}
In recent years, generative models have flourished in deep learning literature, which can generate realistic and high-quality image samples. 
The main objectives of generative models are to estimate the distribution of the training data and to generate samples that are likely to belong to the distribution.
In the image domain, generated images are highly realistic thanks to state-of-the-art deep generative models including variational autoencoders (VAEs)~\cite{vae}, generative adversarial networks (GANs)~\cite{biggan}, and diffusion models~\cite{diffusion}.

Evaluation of generative models is one of the most crucial parts to develop the models.
A classical approach is to measure the likelihood of the real data under the estimated distribution.
However, for intractable models such as GANs, the likelihood cannot be directly computed and estimation of the likelihood is impractical when generated samples are high-dimensional data.
Moreover, \citeauthor{theis2016note}~\shortcite{theis2016note} argued that the likelihood is often unrelated to the quality of generated samples.

Various evaluation metrics have been proposed so far other than the likelihood measure.
In general, they are based on the comparison between the learned distribution and the ground truth distribution in certain feature spaces.
Most early metrics use classification models to embed images into informative feature spaces~\cite{IS,FID,PR,KID}.
For instance, the \Frechet Inception distance (FID)~\cite{FID}, which is the most ubiquitously used metric to this day, uses the Inception model~\cite{Inception} trained on the ImageNet dataset~\cite{imagenet}.
These metrics take advantage of the classification models that are regarded as effective to capture high-level semantic information.


\begin{figure}[t]
     \centering
     \includegraphics[width=\columnwidth]{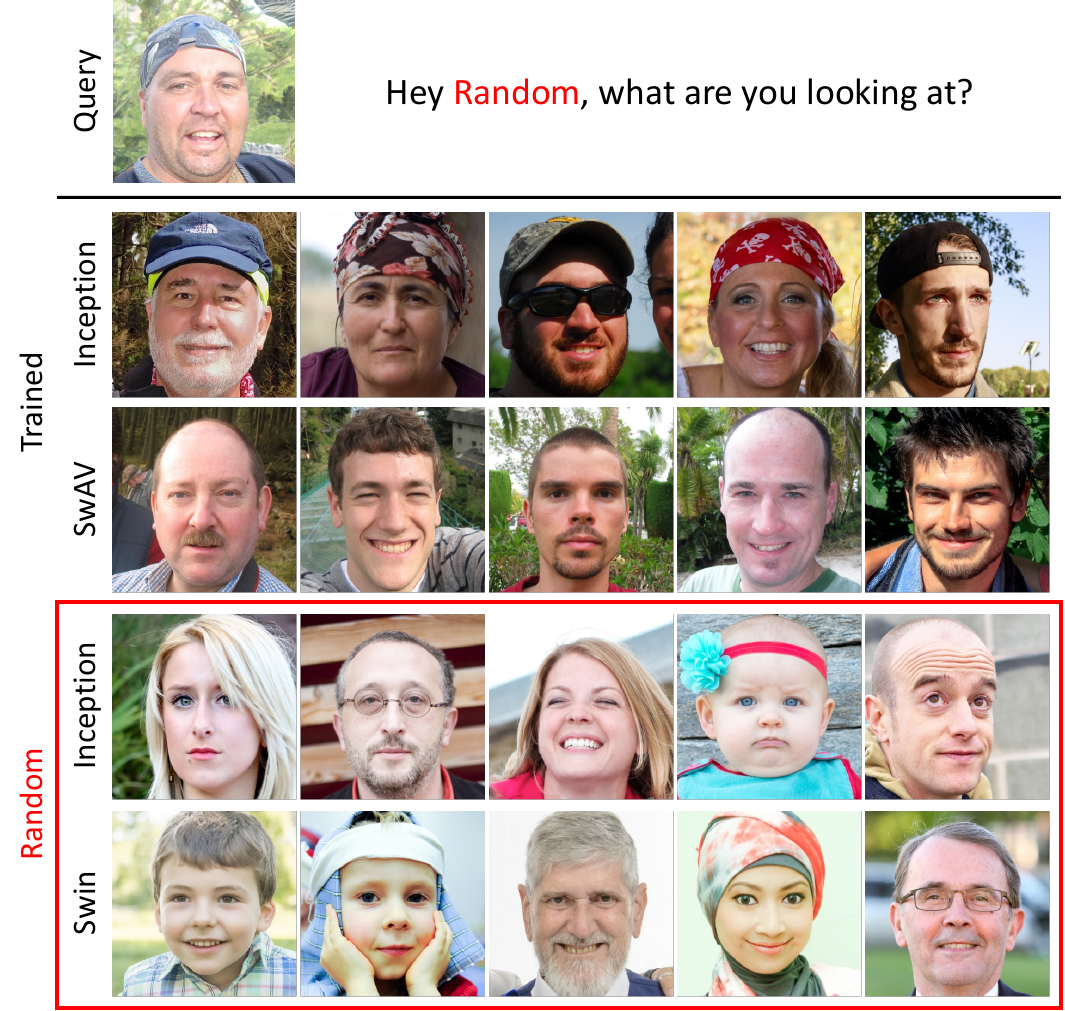}
     \caption{Five nearest neighbors among the ground truth real images by each model for a generated query image. We argue that the random features have different perspectives from the trained ones.} 
    \label{fig:nn}
\end{figure}

However, recent studies have argued that features optimized for classification may be limited for evaluating generative models.
In~\cite{DC}, it was shown that the evaluation results can be underrated in domains different from that of natural images.
It was also observed that the Inception feature yields unreliable evaluation results when it comes to non-ImageNet datasets~\cite{fidself}.
\citeauthor{role}~\shortcite{role} showed that the Inception feature has the null space where the FID can be improved without enhancing the quality of generated images, which is due to the bias of the classification task.

One approach to address these issues is to use the models that are more sophisticated than the classification models.
\citeauthor{fidself}~\shortcite{fidself} proposed to use features from self-supervised learning-based models such as SwAV~\cite{swav} to deal with the images that do not belong to the ImageNet classes such as face, bedroom, and church.
\citeauthor{role}~\shortcite{role} showed that the evaluation results agree with human assessment when the CLIP model~\cite{clip} trained on the vision-language modeling task is employed.

Curiously, randomly initialized networks have been considered as another approach.
In~\cite{DC}, features from convolutional neural networks (CNNs) with random weights yielded sensible evaluation results for the generative models in domains other than natural images such as hand-written digits and speech data.
\citeauthor{sgxl}~\shortcite{sgxl} also employed a randomly initialized CNN to evaluate generative models.
Although features from randomly initialized networks (referred to as ``random features'' in this paper) are occasionally considered in the literature, we notice that justification of using the random features has been insufficiently addressed especially for the domains where the sophisticated features already exist (e.g., natural images in~\figurename~\ref{fig:nn}).  
In particular, the following questions remain unanswered: 
\begin{enumerate}
\item Do random features even capture any useful information?
\item What are the advantages of random features than those from trained networks?
\item Under what circumstances do random features work? 
\end{enumerate}



In this paper, we demystify randomly initialized networks for evaluating generative models via extensive analysis.
To answer the above research questions, first, we examine the feature spaces established by randomly initialized networks by inspecting both sample-based and distribution-based distances. 
Then, we perform in-depth analysis of faulty images created by generative models.
We also demonstrate that random features have distinct perspectives compared to the features from trained networks (referred to as ``trained features'').  
Finally, we present comprehensive evaluation results of various generative models including GANs, VAEs, and diffusion models using random features.


Our main contributions are summarized as follows:
\begin{itemize}
    \item 
    We provide various analyses on the feature spaces of randomly initialized networks in the viewpoint of evaluating generative models.
    Based on the analyses, we establish a fundamental basis of the random features to be employed to evaluate generative models. 
    \item 
    We show the cases where the random features have different characteristics compared to the trained ones.
    We examine the faulty images from generative models to understand how the images appear in the feature spaces.
    It is shown that random features have different criteria to judge the faulty images from the trained features.
    \item 
    We rigorously evaluate generative models with features from various networks with random weights. 
    As a result, we demonstrate that the random features are useful for evaluation of generative models and, furthermore, they can provide information complementary to that by the trained features regarding the characteristics of the generated images.
    
\end{itemize}






\section{Related Work}
\subsection{Generative Models}
Generative models aim to estimate the distribution of the target training data.
Recently, the state-of-the-art models can generate highly realistic images after they are trained on challenging real-world image datasets~\cite{imagenet,sg}.
As one of the likelihood-based generative models, VAEs estimate the distribution by mapping images to the learned latent space~\cite{vae}, but often generate blurry images~\cite{introvae}.
GANs optimize the adversarial loss, which play a min-max game between the generator and discriminator~\cite{goodfellow2014generative}.
GANs significantly improve the visual quality of the generated images, while the estimated densities are intractable.
Diffusion models learn to revert the progressive diffusion process that adds noise to an image.
Recently, they yield comparable results to GANs with tractable estimated densities~\cite{diffusion}.

\subsection{Evaluation of Generative Models}
For likelihood-based generative models, the likelihood measure can be directly adopted for evaluation.
However, it is not the case for GANs since they are based on likelihood-free training.
For GANs, kernel density estimation using Parzen window can be employed as an indirect likelihood measure, but a significantly large number of samples are required for high-resolution images~\cite{theis2016note}.
It was also argued that the likelihood measure is sometimes irrelevant to the quality of samples.

To properly consider the sample quality in an efficient manner, the Inception score was proposed~\cite{IS}, which utilizes the trained classification model, i.e., Inception~\cite{Inception}.
However, since the target distribution is not considered, the evaluation results often fail to detect mode collapse and favor highly classifiable images rather than high quality images.

FID is the \Frechet distance (also known as the 2-Wasserstein distance and the earth mover's distance) between the trained and target distributions in the feature space of the Inception model.
By assuming that the features follow normal distributions, it is computed as follows,
\begin{equation}
    \mathrm{FID}= \left\|\mu^{g}-\mu^{r}\right\|^2_2 + \mathrm{Trace}\left(\Sigma^g+\Sigma^r-2\left(\Sigma^g\Sigma^r\right)^\frac{1}{2}\right), 
\end{equation}
where 
$\left(\mu^g,\Sigma^g\right)$ and $\left(\mu^r,\Sigma^r\right)$ are the sample mean and the covariance of the generated and real data, respectively, in the Inception feature space.
Thanks to its simplicity and reasonable accuracy, FID is ubiquitously used for evaluation of various generative models.
However, limitations of FID have been also pointed out: FID is based on an over-simplified normality assumption~\cite{KID,trend}, has a bias to the number of samples~\cite{infinity}, and is unexpectedly sensitive to subtle pre-processing methods~\cite{aliased}.


Another approach is to separately consider two different aspects of performance (i.e., fidelity and diversity) rather than a single value such as FID~\cite{PR,IPR,DC}.
Such twofold metrics are particularly useful for diagnostic purposes.
There also exist attempts to evaluate the quality of individual samples~\cite{IPR,rarity}.
They are usually based on the distance of an image to its nearest neighbor among a set of real or generated images.

Most of the aforementioned metrics take advantage of classification models (e.g., the Inception model) to obtain meaningful features.
However, it has been argued that the feature spaces are classification-oriented and may not be necessarily perceptual~\cite{fidself,role}.
For alternative feature spaces, models trained on sophisticated tasks such as self-supervised learning and vision-language modeling are employed.
It has been shown that features from the SwAV~\cite{swav} and CLIP~\cite{clip} models are effective for non-ImageNet datasets such as the FFHQ dataset~\cite{sg}.

\subsection{Randomly Initialized Networks}
In order to explain the expressive power of deep neural networks, models with random weights are often considered.
\citeauthor{dip}~\shortcite{dip} showed that the structure of CNNs itself has an implicit prior on the image domain, thus models can perform image super-resolution, in-painting, and de-noising tasks without a large number of training images.
The lottery ticket hypothesis~\cite{lottery, Ramanujan_2020_CVPR} demonstrates that a randomly initialized network contains subnetworks that perform comparably to the trained networks.
Furthermore, features from randomly initialized networks can be used to predict class labels of images~\cite{nrf}.


\citeauthor{DC}~\shortcite{DC} argued that features from the VGG models~\cite{vgg} with random weights are effective for evaluating generated samples in domains other than natural images such as hand-written digits and speech.
In addition, they observed that the random features yield similar evaluation results to the trained features for natural images.
In~\cite{sgxl}, it was also observed that the evaluation results using the feature of the Inception model with  random weights are similar to those using the trained features.
However, we argue that clear grounds beyond these observations are required to rely on random features for evaluating generative models.


\section{Experimental Setup}
%
For the real-world image datasets, we use the ImageNet~\cite{imagenet} and FFHQ~\cite{sg} datasets.
We also use a non-natural cartoon image dataset that consists of \pokemon images. 
For generated images, we use various generative models including E-VDVAE~\cite{vae}, BigGAN~\cite{biggan}, ProjectedGAN~\cite{projgan}, StyleGAN~\cite{sg}, StyleGAN2~\cite{sg2}, StyleGAN-XL~\cite{sgxl}, and ADM~\cite{diffusion}.
For the diffusion models, we use the classification-guidance model (ADM-C) and the up-scaling model (ADM-U).

We use 50,000 generated images for evaluating each generative model. 
We use the official codes of the models with pre-trained weights.
For the ImageNet class-conditional models (i.e., BigGAN, ADM-C, and ADM-U), we generate 50 images for each of the 1,000 classes.
BigGAN and ADM-U generate images at a resolution of 256$\times$256 pixels, while E-VDVAE and ADM-C generate images at a resolution of 64$\times$64 pixels.
For FFHQ, we use StyleGAN and StyleGAN2 at 1024$\times$1024 pixels, and E-VDVAE and ProjectedGAN at 256$\times$256 pixels.
We use ProjectedGAN and StyleGAN-XL trained on \pokemon to generate images at a resolution of 256$\times$256 pixels.

For trained features, the Inception model~\cite{Inception} and the SwAV model with the ResNet50 structure~\cite{swav} are used.
For random features, we employ the Inception and VGG~\cite{vgg} models that have been used in the previous studies~\cite{sgxl,DC}.
We also consider Transformer models: Swin-T~\cite{swin} and ViT-T~\cite{vit} with random weights.
The Kaiming uniform~\cite{init} is used to initialize the weights.
The evaluation results using the random features are averaged over five runs using different seeds for the random state initialization.
Variations across the five runs are provided in the Supplementary Material, where we do not observe significant difference between runs.

For evaluation metrics, we employ the FID using various trained and random models for feature embedding.
We provide further results using different metrics including twofold metrics in the Supplementary Material, which are consistent to the results using FID.


\section{Understanding Random Feature Spaces}

In general, the evaluation methods of generative models are based on the comparison between the learned distribution and the actual distribution in certain feature spaces.
Therefore, understanding the feature spaces is crucial to properly interpret the evaluation results.
In this section, we examine the feature spaces of randomly initialized networks through comparison with trained feature spaces: 1) which information they capture, and 2) how well they differentiate disturbances.

\subsection{Nearest Neighbor Retrieval}
First, we examine the characteristics of the images that are closely located in a feature space. 
We choose a query image generated from StyleGAN2 trained on FFHQ and retrieve its nearest real samples with respect to trained or random feature spaces.
We use Inception and SwAV to extract trained features, and Inception and Swin with random weights to extract random ones.


\figurename~\ref{fig:nn} shows a query image and its five nearest images with respect to each feature space.
In the case of trained Inception, all retrieved samples contain adults wearing caps, which is one of the classes in the ImageNet dataset.
SwAV retrieves images of adult males in front of green foliage as in the query image even if they do not wear caps.
In the cases of both random features, the retrieved samples are overall in similar colors and brightness to the query image but highly diverse in terms of age, gender, background, etc.
In summary, we observe that while the trained features mostly focus on high-level semantic information such as gender, age, backgrounds, and accessories, the random features mostly focus on low-level information such as color and brightness.

\subsection{Disturbance}

Next, we examine how well random features capture differences between distributions.
We apply various low-level image disturbances such as blur, noise, and color jitter to a set of images. 
We also consider the disturbance in terms of high-level semantics, called class contamination, which replaces some images with those from different datasets.
Then, we measure the difference between the original and disturbed sets using FID, but with randomly initialized Inception, VGG, Swin, and ViT as feature extractors.
For comparison, the trained Inception and SwAV are also employed.

\begin{figure}[t]
     \centering
     \includegraphics[width=0.85\columnwidth]{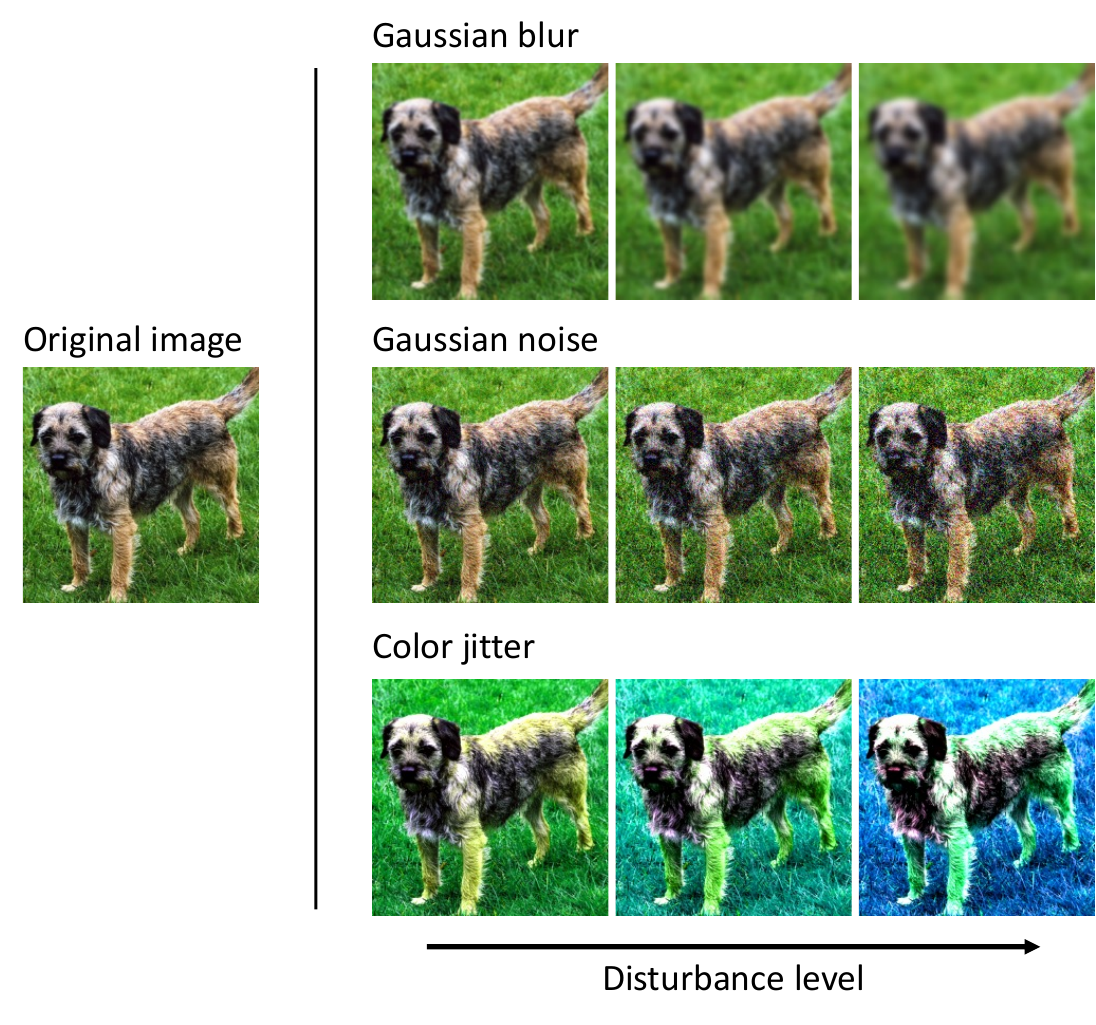}
     \caption{Example of disturbed images}
    \label{fig:disturb}
\end{figure}

For the  50,000 images from the ImageNet validation dataset, three increasing levels of disturbances are applied (\figurename~\ref{fig:disturb}): Gaussian blur with width $\sigma_{gb} \in \{1,2,3\}$, Gaussian noise with variance $\sigma_{gn}^2 \in \{0.05, 0.10, 0.15\}$, color jitter that changes all of the brightness, contrast, saturation, and hue with ratio $r_c \in \{0.1, 0.2, 0.3\}$, and class contamination with replacement ratio $r_r \in \{0.25, 0.5, 0.75\}$.

\begin{table}[t]
\centering
\scriptsize
\begin{tabular}{cccccc}
\toprule
Feature &  Level & \specialcell{Gaussian\\blur} & \specialcell{Gaussian\\noise} & \specialcell{Color\\ jitter} & \specialcell{Class\\contamination}\\
\midrule
\multirow{3}{*}{\specialcell{Inception\\(trained)}}    
                        & 1                 & 7.81          & 7.95         & 17.29 & 19.64\\
                      & 2                 & 23.13          & 21.34        & 22.93 & 50.14\\
                      & 3                 & 40.67          & 41.09        & 28.16 & 89.63\\
\midrule
\multirow{3}{*}{\specialcell{SwAV\\(trained)}}   
                        & 1                 & 0.21           & 0.64       & 6.62 & 3.70\\
                      & 2                 & 1.45          & 1.92         & 7.02 & 9.50\\
                      & 3                 & 10.66          & 3.41        & 7.37 & 17.50\\
\midrule
\multirow{3}{*}{\specialcell{Inception\\(random)}}
                        & 1                 & 29.88         & 0.10         & 469.89 & 2.22\\
                      & 2                 & 92.84         & 1.05          & 482.55 & 8.79\\
                      & 3                 & 148.37        & 4.00          & 496.70 & 20.86\\
\midrule
\multirow{3}{*}{\specialcell{VGG\\(random)}} 
                        & 1                &0.20    &$<$0.01&4.14   &0.03\\
                      & 2                 &0.54    &0.02   &4.29   &0.09\\
                      & 3                 &0.84    &0.07   &4.46   &0.20\\
\midrule
\multirow{3}{*}{\specialcell{Swin\\(random)}} 
                        & 1                 & 0.54         & $<$0.01     & 436.09 & 3.66\\
                      & 2                 & 1.77         & 0.04         & 449.63 & 14.34\\
                      & 3                 & 3.05         & 0.16         & 459.22 & 32.77\\
\midrule
\multirow{3}{*}{\specialcell{ViT\\(random)}}
                        & 1                 &0.17   &$<$0.01&220.25 &2.32\\
                      & 2                 &0.63    &0.01   &225.77 &8.86\\
                      & 3                 &1.20    &0.04   &229.07 &19.90\\
\bottomrule
\end{tabular}
\caption{FID of disturbed images using features from trained and random models. The values for Inception (random) are scaled by $\times$10\textsuperscript{14}.}
\label{tab:disturbance}
\end{table}

\tablename~\ref{tab:disturbance} shows FID values between the original and disturbed images.
Overall, all features correctly yield larger FID along increasing levels of disturbances.
However, the trained and random features differently assess the distributions of the disturbed images with respect to the type of disturbances.
For instance, both trained Inception and SwAV determine that the class contamination at level 3 causes the largest difference in the distribution.
On the other hand, all random features give the largest score to the color jitter at level 3.
These results are consistent with the observations in \figurename~\ref{fig:nn}, i.e., the trained features mostly focus on the high-level semantics such as class labels.

In terms of low-level disturbances, the trained Inception feature judges the degradations due to the noise, blur, and color jitter to be similar at level 2.
SwAV also considers the blurred and noisy images to have similar scores at level 2. 
Considering that the noise is less perceptible for humans than the others (\figurename~\ref{fig:disturb}), it can be argued that the trained Inception and SwAV features have excessive sensitivity to noise.  
On the contrary, all of the random features judge that the noise and color jitter cause the smallest and the largest difference, respectively, which is consistent with human perception for the low-level disturbances. 




Therefore, 
the trained models and randomly initialized models have different criteria to judge how the distributions become different under disturbances.
In most cases, the random features are more precise to the changes in low-level information such as color, blur, and noise.
On the other hand, the trained features focus on high-level semantics such as class categories, and thus are often inconsistent with the human perception in terms of low-level information.

\section{Disturbances of Generated Images}
In the previous section, we discovered clues on how the different feature models regard differences in distributions using the disturbances applied to real images in a controlled manner.
In this section, we investigate the disturbances that practically occur in generative models, in the viewpoint of how they appear in the trained or random feature spaces.

\subsection{Can We Distinguish Disturbances in Generated Images?}
In order to determine faulty images, we use a method similar to the outlier detection method~\cite{DC} with modifications in the criteria.
We consider two feature spaces to separate different types of outliers.
First, we use the pre-trained CLIP model that is designed to extract high-level semantic information for vision-language relationship modeling~\cite{clip}.
We employ the visual encoder part of the CLIP model with a vision transformer structure. 
Second, we consider the lower stage of the CLIP model to develop a feature space that focuses on low-level information.
Specifically, we use the output of the first convolutional layer of CLIP after global average pooling, which is also known as the convolutional stem.

To detect outliers, we compute the nearest neighbor distance $d_\mathtt{m}$ for a set of generated images, which is the distance between the feature of an image and that of its $k$\textsuperscript{th} nearest neighbor ($k=5$), where the features are the outputs of a model $\mathtt{m} \in \{\mathtt{CLIP},\mathtt{STEM}\}$.
We compute $d_\mathtt{CLIP}$ and $d_\mathtt{STEM}$, which focus more on high-level and low-level information, respectively.
Then, we obtain a set of outliers such that they have the largest $d_\mathtt{m}$.
In order to separate the high-level and low-level outliers, we consider the images that belong to only a single set of outliers and disregard outliers contained in both sets.
More specifically, we take the images that have the top $\alpha$\% largest $d_\mathtt{CLIP}$ and the bottom $(100-\alpha)$\% largest $d_\mathtt{STEM}$ for high-level outliers, and vice versa for low-level outliers. 
As a result, the outliers are far from the other samples, i.e., the corresponding feature space regards them as having distinct characteristics that are rare in the others.

\begin{figure}[t]
     \centering
     \begin{subfigure}[b]{\linewidth}
         \centering
         \includegraphics[width=0.9\textwidth]{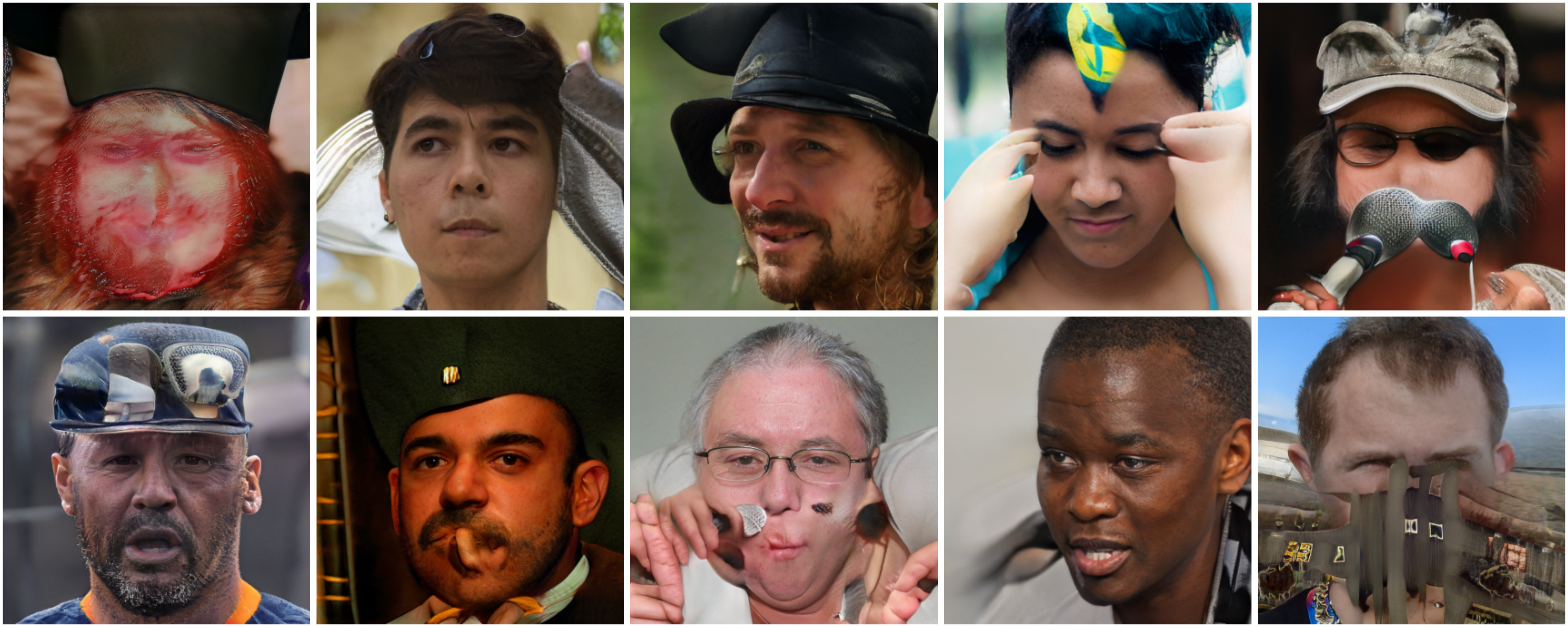}
         \caption{High-level outliers}
         \label{fig:sg2-high}
     \end{subfigure}
     \begin{subfigure}[b]{\linewidth}
         \centering
         \includegraphics[width=0.9\textwidth]{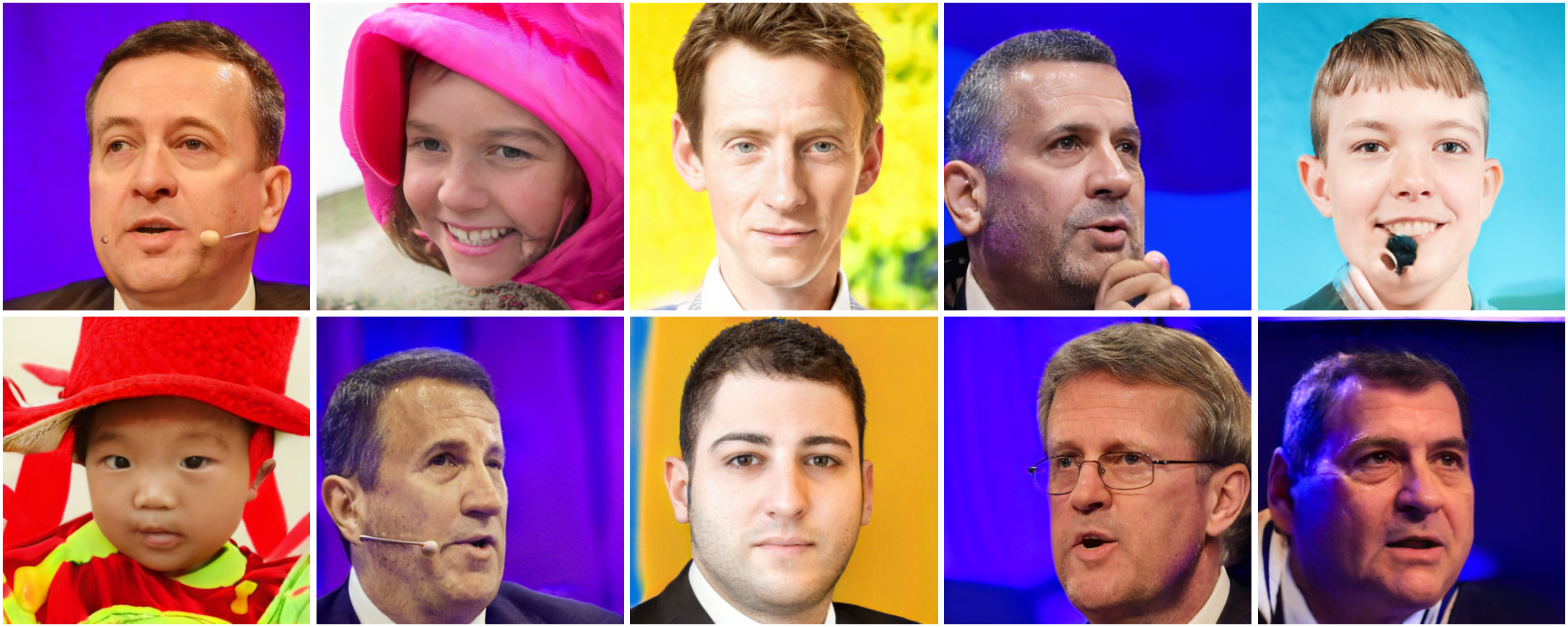}
         \caption{Low-level outliers}
         \label{fig:sg2-low}
     \end{subfigure}
        \caption{StyleGAN2 outliers identified using $d_\mathtt{CLIP}$ and $d_\mathtt{STEM}$}
        \label{fig:sg2-outliers}
\end{figure}

\figurename~\ref{fig:sg2-outliers} shows the high-level and low-level outliers generated by StyleGAN2 trained on FFHQ ($\alpha$=67\%). 
In \figurename~\ref{fig:sg2-high}, the images have severe errors in high-level semantics such as unidentifiable faces, body parts, and accessories.
On the other hand, in \figurename~\ref{fig:sg2-low}, most of the images show acceptable quality of faces but with vivid colors around them.
Outliers for other datasets are presented in the Supplementary Material.
The evaluation metrics would regard both outliers as deviated compared to the real FFHQ dataset.
The high-level outliers are obviously failure cases for modeling the real distribution, whereas the low-level outliers are rare in the real distribution, which is also an abnormality in a different sense.
Therefore, it would be desirable to correctly consider these outliers when evaluating generative models, which is examined below.

\subsection{How Do the Disturbances of Outliers Appear in Different Feature Spaces?}
We analyze how a feature space regards changes in distributions by the different types of outliers.
First, we build two subsets containing high- and low-level outlier images, respectively, from a set of images synthesized by a generative model trained on a real dataset.
Then, we progressively replace ten images from the real dataset with the outliers from one of the subsets.
For each step of replacement, we measure the FID between the original real dataset and the replaced set in various feature spaces.
In this case, the distributions of the replaced sets can be regarded as distributions estimated by faulty generative models and the number of replaced images as the degree of the difference of the distributions.

\begin{figure}[t]
     \centering
     \begin{subfigure}[b]{0.36\linewidth}
         \centering
         \includegraphics[width=\textwidth]{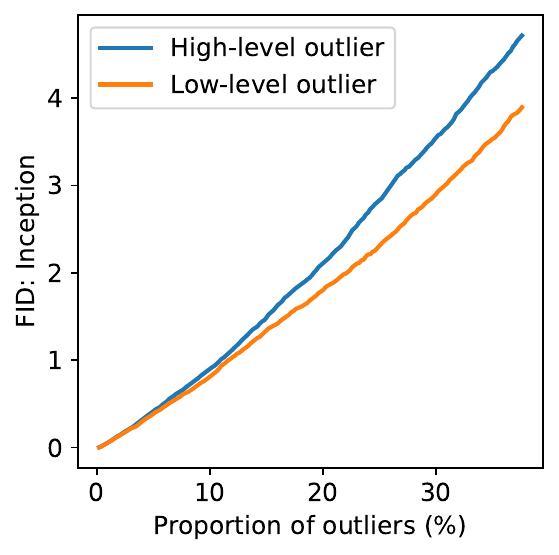}
         \caption{Inception}
         \label{fig:replace-inception}
     \end{subfigure}
     \hspace{1.8mm}
     \begin{subfigure}[b]{0.39\linewidth}
         \centering
         \includegraphics[width=\textwidth]{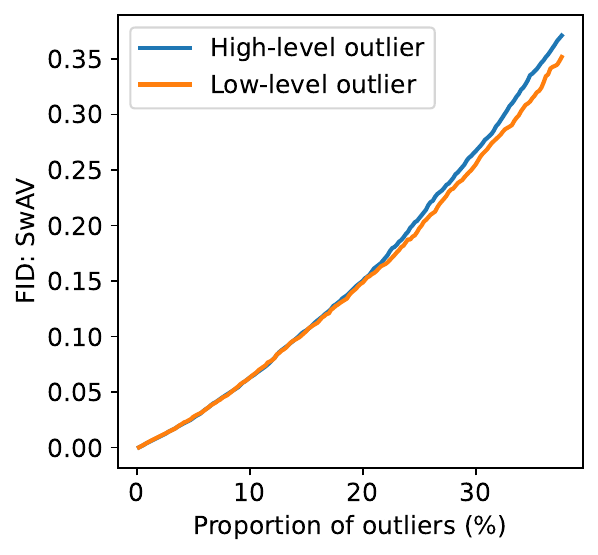}
         \caption{SwAV}
         \label{fig:replace-swav}
     \end{subfigure}
     \begin{subfigure}[b]{0.36\linewidth}
         \centering
         \includegraphics[width=\textwidth]{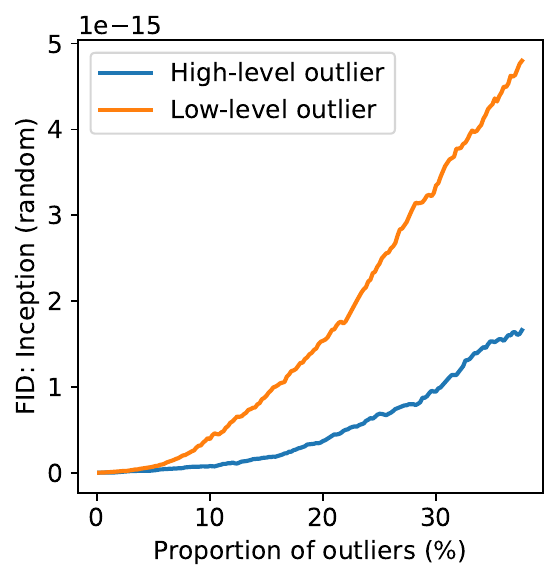}
         \caption{Inception (random)}
         \label{fig:replace-inception-r}
     \end{subfigure}
     \hspace{1.6mm}
     \begin{subfigure}[b]{0.40\linewidth}
         \centering
         \includegraphics[width=\textwidth]{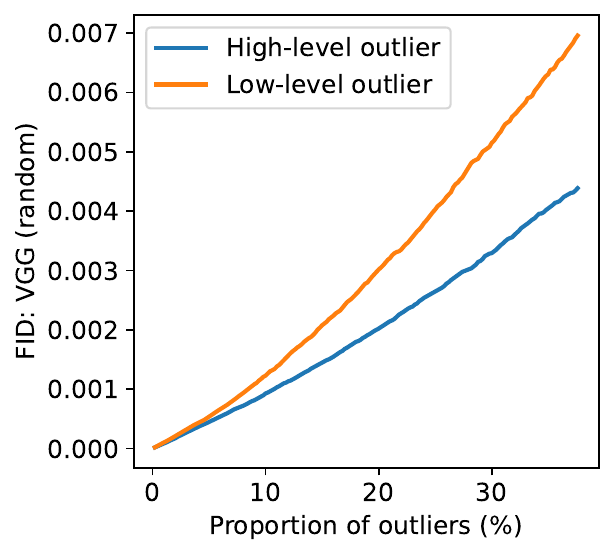}
         \caption{VGG (random)}
         \label{fig:replace-vgg-r}
     \end{subfigure}
     \begin{subfigure}[b]{0.38\linewidth}
         \centering
         \includegraphics[width=\textwidth]{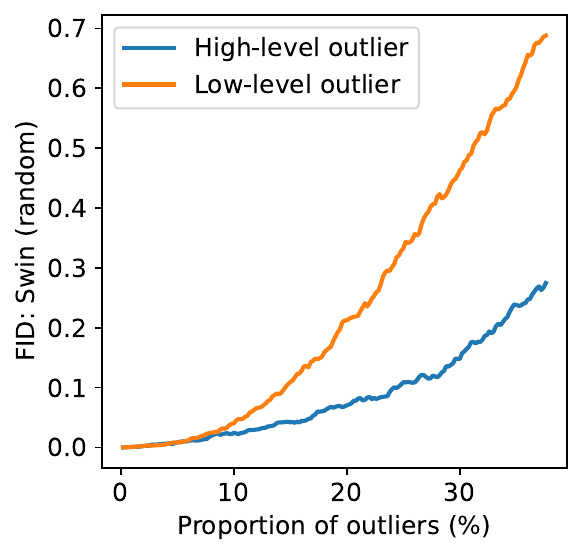}
         \caption{Swin (random)}
         \label{fig:replace-swin-r}
     \end{subfigure}
     \hspace{2.2mm}
     \begin{subfigure}[b]{0.38\linewidth}
         \centering
         \includegraphics[width=\textwidth]{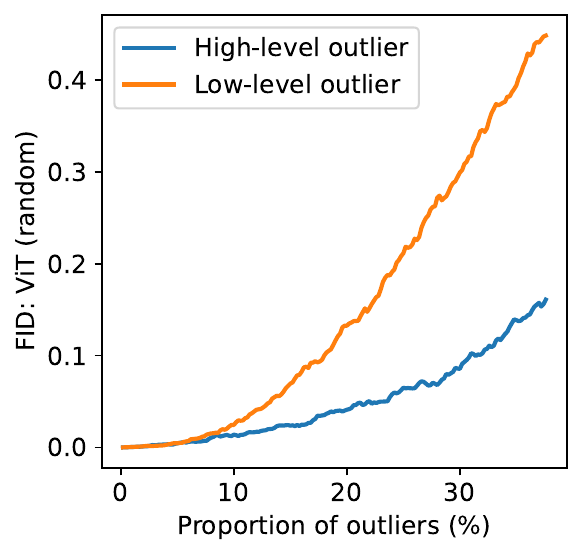}
         \caption{ViT (random)}
         \label{fig:replace-vit-r}
     \end{subfigure}
     \hspace{0.4mm}
        \caption{FID vs. the proportion of outliers for different features}
        \label{fig:replace}
\end{figure}

In this experiment, we randomly sample 5,000 images from FFHQ as a real dataset.
Among a set of 10,000 images generated by StyleGAN2, we compose two subsets consisting of 1,880 high-level and low-level outlier images, respectively.
We only replace real images at each step, while keeping the previously added outliers. 
For feature spaces, we use two trained models (Inception and SwAV) and four randomly initialized models (Inception, VGG, Swin, and ViT).
\figurename~\ref{fig:replace} shows the FID with respect to the proportion of the outliers in different feature spaces.

In \figurename s~\ref{fig:replace-inception} and~\ref{fig:replace-swav}, replacing with high-level outliers yields larger FID than replacing with low-level outliers, indicating that both models judge the high-level outliers to incur larger difference.
Still, both trained features are quite sensitive to the low-level outliers by assessing them to be comparable to the high-level outliers. 
On the contrary, all random features regard that the low-level outliers change the distribution much more than the high-level outliers (\figurename s~\ref{fig:replace-inception-r}-\ref{fig:replace-vit-r}).
Note that the random features tend to fluctuate more along gradual changes in the distribution, which might be risky when distributions with subtle differences are compared.
Results of different datasets are shown in the Supplementary Material.
These results indicate that sensitivity to different types of outliers is different depending on whether the feature is learned or random, which are consistent to those in \tablename~\ref{tab:disturbance}, especially for class contamination and color jitter.
The trained features are more sensitive to the high-level outliers than the low-level outliers, which also assess class contamination as severer degradation than color jitter.
On the other hand, the random features are sensitive to the low-level outliers, which yield the largest FID for color jitter.

\section{Do Random Features Correctly Evaluate Generative Models?}

\subsection{Comprehensive Evaluation}

\begin{table}[t]
\centering
\scriptsize
\begin{tabular}{@{}lrrrrrr@{}}
\toprule
\textbf{Datasets} & \multicolumn{2}{c}{Trained} & \multicolumn{4}{c}{Random} \\ \cmidrule(l){2-3} \cmidrule(l){4-7}
Models & Inception & SwAV & Inception & VGG & Swin & ViT \\ \midrule
\textbf{ImageNet $256^2$}&        &      & \multicolumn{2}{l}{\hspace{4.5mm}$\times$10\textsuperscript{-15}}\\
BigGAN            & 6.24      & 4.45     &  2.73        & 0.009   &  1.07 & 0.62    \\
ADM-U             & 4.94      & 2.25     &  8.52        & 0.012   &  3.57 & 1.61    \\ \midrule
\textbf{ImageNet $64^2$}&         &      &              &    &              \\
E-VDVAE           & 46.23     & 6.52     &  367.68      & 0.279   &  4.58 &2.17       \\
ADM-C             & 8.91      & 1.24     &  28.95       & 0.021   &  2.92 &1.23       \\ \midrule
\textbf{FFHQ $1024^2$}&           &      &              &    &              \\
StyleGAN          & 4.58      & 0.92     &  4.62        & 0.008   &  0.34 & 0.18      \\
StyleGAN2         & 3.10      & 0.42     &  0.49        & 0.005   &  0.23 & 0.13      \\ \midrule
\textbf{FFHQ $256^2$} &           &      &              &    &              \\
E-VDVAE           & 72.10     & 3.42     &  321.11      & 0.447   &  7.87 & 4.87      \\
ProjectedGAN      & 4.18      & 1.32     &  18.01       & 0.028   &  0.68 & 0.35      \\ \midrule
\textbf{\pokemon $256^2$}  &           &      &         &         &              \\
ProjectedGAN      & 27.88     & 4.80     &  90.40        & 0.343   &  3.86 & 2.79      \\
StyleGAN-XL       & 25.46     & 3.43     &  31.22        & 0.271   &  1.12 & 0.67      \\
\bottomrule
\end{tabular}
\caption{Evaluation results in terms of FID using different feature spaces for various generative models}
\label{tab:eval-FID} 
\end{table}

\begin{figure}[t]
     \centering
     \includegraphics[width=0.85\columnwidth]{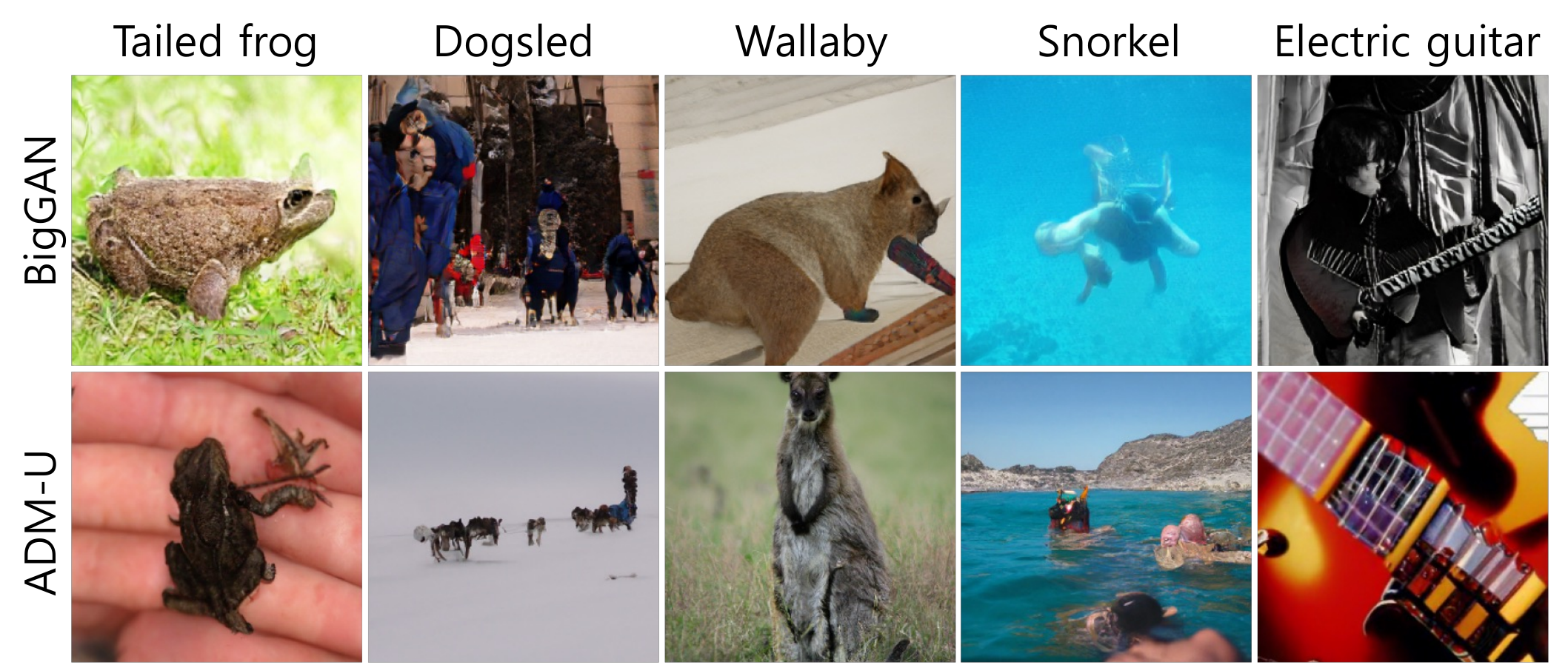}
     \caption{Generated images of BigGAN and ADM-U trained on ImageNet. The samples are randomly selected.}
    \label{fig:imagenet}
\end{figure}

In this section, we conduct comprehensive evaluation of various generative models using random features.
\tablename~\ref{tab:eval-FID} shows the results using different features for FID.
In most cases, in terms of determining the superiority of one
against the other within a dataset, both trained and random features yield consistent results except for BigGAN and ADM-U.
We observe that while BigGAN and ADM-U can generate decent samples, a considerable number of faulty samples with diverse disturbances are generated as shown in~\figurename~\ref{fig:imagenet}. 
Thus, it would be necessary to utilize various feature spaces that could consider various aspects of faulty samples.
We will discuss more later.

\subsubsection{Do the Results Vary with Different Random Seeds for Initialization?}

\hyphenation{BigGAN}
\hyphenation{StyleGAN}


\begin{figure}[t]
     \centering
     \begin{subfigure}[b]{0.37\linewidth}
         \centering
         \includegraphics[width=\textwidth]{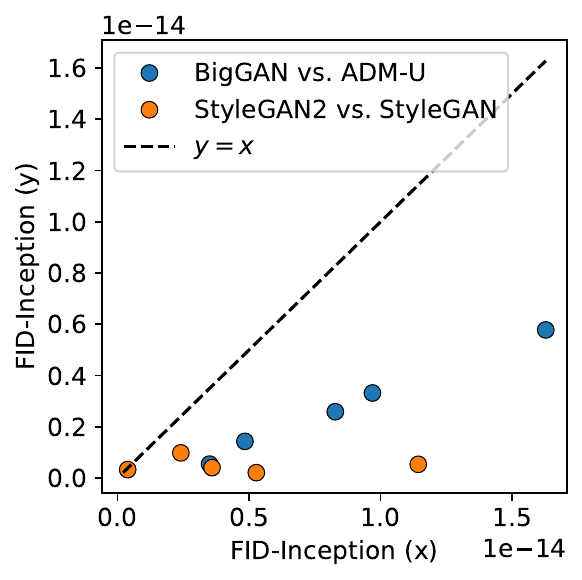}
         \caption{Inception}
         \label{fig:seed-inception}
     \end{subfigure}
    \hspace{1.5mm}
     \begin{subfigure}[b]{0.39\linewidth}
         \centering
         \includegraphics[width=\textwidth]{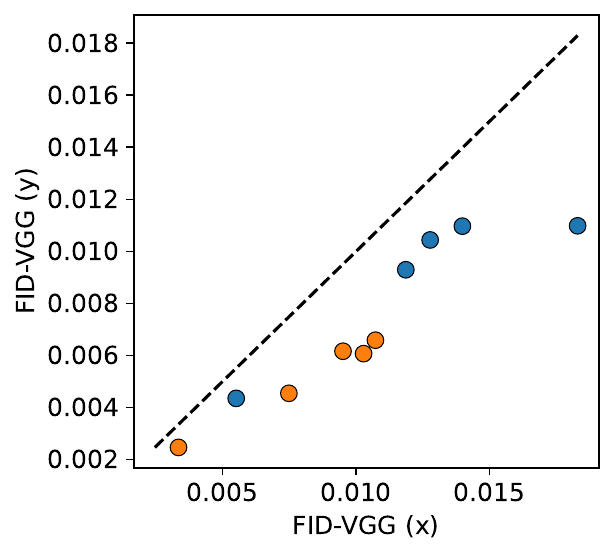}
         \caption{VGG}
         \label{fig:seed-vgg}
     \end{subfigure}
     \begin{subfigure}[b]{0.37\linewidth}
         \centering
         \includegraphics[width=\textwidth]{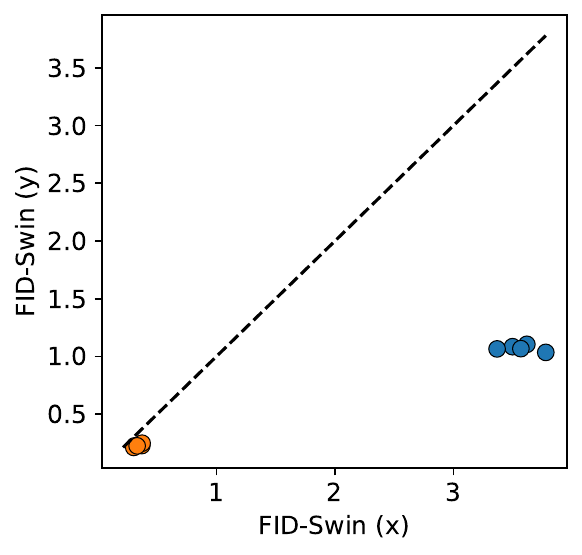}
         \caption{Swin}
         \label{fig:seed-swin}
     \end{subfigure}
    \hspace{2mm}
     \begin{subfigure}[b]{0.37\linewidth}
         \centering
         \includegraphics[width=\textwidth]{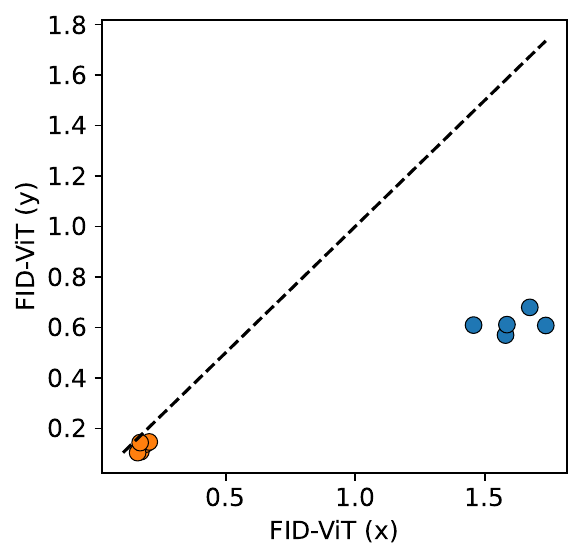}
         \caption{ViT}
         \label{fig:seed-vit}
     \end{subfigure}
        \caption{Evaluation results of generative models using random features with five different seeds for weight initialization of the feature networks. The blue circles are FIDs of ADM-U (x-axis) and BigGAN (y-axis), and the orange ones are FIDs of StyleGAN (x-axis) and StyleGAN2 (y-axis).}
        \label{fig:seed}
\end{figure}

We examine how much the evaluation results using random features vary along different random seeds for weight initialization in \figurename~\ref{fig:seed}. 
We can observe that all data points are below the dashed lines, which indicates that the results of determining superior models are not reversed depending on the seed.
In \figurename s~\ref{fig:seed-inception} and~\ref{fig:seed-vgg}, FIDs of the CNNs with random weights vary with random seeds. 
Surprisingly, FIDs using Swin and ViT are highly consistent over the five different seeds as shown in \figurename s~\ref{fig:seed-swin} and \ref{fig:seed-vit}.
We surmise that successive softmax operations in the Transformer blocks are helpful in this consistency because they normalize intermediate features.
Consequently, in order to obtain reliable results, it would be better to employ a Transformer structure and to incorporate the results using different random seeds (e.g., by averaging).

\subsubsection{Do the Results Vary with Different Model Sizes?}

\begin{table}[t]
\centering
\scriptsize
\begin{tabular}{@{}lccccc@{}}
\toprule
Models & \alignc{Swin-T} & \alignc{Swin-S} & \alignc{Swin-B} & \alignc{ViT-T}& \alignc{ViT-B} \\ 
\#Parameters & 28M & 50M & 88M & 6M & 86M \\ \midrule
\textbf{ImageNet $256^2$} & \\
BigGAN & 1.07 & 1.07 & 1.54 & 0.62 & 3.14\\
ADM-U & 3.57 & 3.70 & 5.03 & 1.61 & 6.61\\
\bottomrule
\end{tabular}
\caption{FIDs of BigGAN and ADM-U using random features from Swin and ViT with different model sizes}
\label{tab:structure} 
\end{table}

In addition, we investigate whether the evaluation results are affected by the size of randomly initialized models.
Based on the results in \figurename~\ref{fig:seed}, we consider Swin and ViT models that show consistent results over the different random seeds.
We evaluate BigGAN and ADM-U (trained on ImageNet $256^2$) with random features from larger sizes of Swin and ViT (i.e., Swin-S, Swin-B, and ViT-B).
Swin-S has more Transformer blocks in an intermediate layer compared to Swin-T.
Swin-B produces higher dimensional outputs than the others (768$\rightarrow$1024) with the same number of blocks to that of Swin-S.
ViT-B has a deeper structure with a higher dimension of outputs (192$\rightarrow$768) than ViT-T.

\tablename~\ref{tab:structure} shows the results. 
Swin-T and Swin-S show similar FIDs, and the superiority of BigGAN is also consistent in Swin-B although the scale of FID increases.
ViT-B is also consistent to ViT-T in determining superiority of BigGAN to ADM-U except for the increase of the FID scale.
Overall, changing the depth of Swin hardly affects the evaluation results and the dimension of the features of Swin and ViT is related to the scale of FID. 
Thus, any of the tested models can be used for evaluation, and smaller models like Swin-T and ViT-T have the advantage of computational efficiency.

\subsection{Are Random Features Adoptable?}
In the previous sections, evaluation results using random features are mostly similar to those using trained features.
However, the random features sometimes yield different results from the trained ones (\tablename s~\ref{tab:disturbance} and \ref{tab:eval-FID}).
This is because the random and trained features for the generated images become different, especially for high- and low-level outliers (\figurename~\ref{fig:replace}). 
Thus, we propose to use both types of features in order to evaluate generative models in various aspects.

To examine when the complementarity of the two features can be expected, we measure the Spearman's rank order correlation coefficient between $d_\mathtt{CLIP}$ and $d_\mathtt{STEM}$ for a set of generated images. 
If the correlation is low, the images have different ranks in terms of $d_\mathtt{CLIP}$ and $d_\mathtt{STEM}$.
In other words, high-level and low-level outliers are distinct.
As shown in~\figurename~\ref{fig:replace}, in this case, the evaluation results would be different depending on whether the used feature is trained or random.

For FFHQ, the correlation coefficients of ProjectedGAN, StyleGAN, and StyleGAN2 images are 0.276, 0.238, and 0.152, respectively.
For ImageNet, the correlation coefficients of ADM-U and BigGAN are 0.191 and \textbf{0.127}, respectively.
In the case of BigGAN, which shows the lowest correlation coefficient, random and trained features are especially likely to yield different evaluation results, which is actually observed (\tablename~\ref{tab:eval-FID}).
As shown in \figurename~\ref{fig:imagenet}, it is quite challenging for generative models to learn the distribution of a thousand of classes of the ImageNet dataset.
Thus, it would be more likely for generative models to produce various types of faulty images.

Thus, to consider different aspects of outliers, it would be desirable to use both trained and random features for evaluation.
In addition, according to the results in \figurename~\ref{fig:seed}, we suggest to use a model with a Transformer structure (e.g., Swin or ViT) for random features in order to minimize the effect of randomness.

\section{Conclusion}
In this paper, we demystified randomly initialized networks for evaluating generative models.
We investigated the random features in various aspects as follows.
\begin{itemize}
    \item 
    We performed in-depth analysis of random features by examining how the random features capture sample-based and distribution-based differences. 
    We observed that random features focus more on low-level information (e.g., color and brightness) compared to trained ones that focus on high-level semantics (e.g., class labels).
    \item 
    We investigated faulty images from generative models and separated them into high-level and low-level outliers. 
    We also demonstrated how the differences in the distribution are assessed by random features by replacing real images with the outliers.
    We found that random features are sensitive to the low-level outliers, showing larger FID to the low-level outliers than the high-level outliers.
    \item 
    We rigorously evaluated various state-of-the-art generative models using random features. 
    We found that while the evaluation results are similar to those using trained features in most cases, the random features yield different evaluation results from the trained ones when generated images are distinct in terms of high-level and low-level outliers.
    Therefore, it can be beneficial to employ both trained and random features for thorough evaluation of generative models.
\end{itemize}


\bibliography{aaai23}

\begin{thebibliography}{37}
\providecommand{\natexlab}[1]{#1}

\bibitem[{Amid et~al.(2022)Amid, Anil, Kotłowski, and Warmuth}]{nrf}
Amid, E.; Anil, R.; Kotłowski, W.; and Warmuth, M.~K. 2022.
\newblock Learning from randomly initialized neural network features.
\newblock arXiv:2202.06438.

\bibitem[{Bińkowski et~al.(2018)Bińkowski, Sutherland, Arbel, and
  Gretton}]{KID}
Bińkowski, M.; Sutherland, D.~J.; Arbel, M.; and Gretton, A. 2018.
\newblock Demystifying {MMD} {GAN}s.
\newblock In \emph{Proceedings of the International Conference on Learning
  Representations}.

\bibitem[{Brock, Donahue, and Simonyan(2019)}]{biggan}
Brock, A.; Donahue, J.; and Simonyan, K. 2019.
\newblock Large scale {GAN} training for high fidelity natural image synthesis.
\newblock In \emph{Proceedings of the International Conference on Learning
  Representations}.

\bibitem[{Caron et~al.(2020)Caron, Misra, Mairal, Goyal, Bojanowski, and
  Joulin}]{swav}
Caron, M.; Misra, I.; Mairal, J.; Goyal, P.; Bojanowski, P.; and Joulin, A.
  2020.
\newblock Unsupervised learning of visual features by contrasting cluster
  assignments.
\newblock In \emph{Advances in Neural Information Processing Systems}.

\bibitem[{Chong and Forsyth(2020)}]{infinity}
Chong, M.~J.; and Forsyth, D. 2020.
\newblock Effectively unbiased {FID} and {Inception} score and where to find
  them.
\newblock In \emph{Proceedings of the IEEE Conference on Computer Vision and
  Pattern Recognition}, 6070--6079.

\bibitem[{{Deng} et~al.(2009){Deng}, {Dong}, {Socher}, {Li}, {Kai Li}, and {Li
  Fei-Fei}}]{imagenet}
{Deng}, J.; {Dong}, W.; {Socher}, R.; {Li}, L.-J.; {Kai Li}; and {Li Fei-Fei}.
  2009.
\newblock ImageNet: A large-scale hierarchical image database.
\newblock In \emph{Proceedings of the IEEE Conference on Computer Vision and
  Pattern Recognition}, 248--255.

\bibitem[{Dhariwal and Nichol(2021)}]{diffusion}
Dhariwal, P.; and Nichol, A.~Q. 2021.
\newblock Diffusion models beat {GANs} on image synthesis.
\newblock In \emph{Advances in Neural Information Processing Systems}.

\bibitem[{Dosovitskiy et~al.(2020)Dosovitskiy, Beyer, Kolesnikov, Weissenborn,
  Zhai, Unterthiner, Dehghani, Minderer, Heigold, Gelly et~al.}]{vit}
Dosovitskiy, A.; Beyer, L.; Kolesnikov, A.; Weissenborn, D.; Zhai, X.;
  Unterthiner, T.; Dehghani, M.; Minderer, M.; Heigold, G.; Gelly, S.; et~al.
  2020.
\newblock An image is worth 16x16 words: Transformers for image recognition at
  scale.
\newblock In \emph{Proceedings of the International Conference on Learning
  Representations}.

\bibitem[{Frankle and Carbin(2019)}]{lottery}
Frankle, J.; and Carbin, M. 2019.
\newblock The lottery ticket hypothesis: Finding sparse, trainable neural
  networks.
\newblock In \emph{Proceedings of the International Conference on Learning
  Representations}.

\bibitem[{Goodfellow et~al.(2014)Goodfellow, Pouget-Abadie, Mirza, Xu,
  Warde-Farley, Ozair, Courville, and Bengio}]{goodfellow2014generative}
Goodfellow, I.~J.; Pouget-Abadie, J.; Mirza, M.; Xu, B.; Warde-Farley, D.;
  Ozair, S.; Courville, A.; and Bengio, Y. 2014.
\newblock Generative adversarial nets.
\newblock In \emph{Advances in Neural Information Processing Systems},
  volume~27, 2672--2680.

\bibitem[{Han et~al.(2022)Han, Choi, Choi, Kim, Ha, and Choi}]{rarity}
Han, J.; Choi, H.; Choi, Y.; Kim, J.; Ha, J.-W.; and Choi, J. 2022.
\newblock Rarity score: A new metric to evaluate the uncommonness of
  synthesized images.
\newblock arXiv:2206.08549.

\bibitem[{Hazami, Mama, and Thurairatnam(2022)}]{vae}
Hazami, L.; Mama, R.; and Thurairatnam, R. 2022.
\newblock Efficient-{VDVAE}: Less is more.
\newblock arXiv:2203.13751.

\bibitem[{He et~al.(2015)He, Zhang, Ren, and Sun}]{init}
He, K.; Zhang, X.; Ren, S.; and Sun, J. 2015.
\newblock Delving deep into rectifiers: Surpassing human-level performance on
  ImageNet classification.
\newblock In \emph{Proceedings of the International Conference on Computer
  Vision}, 1026--1034.

\bibitem[{Heusel et~al.(2017)Heusel, Ramsauer, Unterthiner, Nessler, and
  Hochreiter}]{FID}
Heusel, M.; Ramsauer, H.; Unterthiner, T.; Nessler, B.; and Hochreiter, S.
  2017.
\newblock {GAN}s trained by a two time-scale update rule converge to a local
  {Nash} equilibrium.
\newblock In \emph{Advances in Neural Information Processing Systems},
  6629--6640.

\bibitem[{Huang et~al.(2018)Huang, He, Sun, Tan et~al.}]{introvae}
Huang, H.; He, R.; Sun, Z.; Tan, T.; et~al. 2018.
\newblock {IntroVAE}: Introspective variational autoencoders for photographic
  image synthesis.
\newblock In \emph{Advances in Neural Information Processing Systems}.

\bibitem[{Karras, Laine, and Aila(2019)}]{sg}
Karras, T.; Laine, S.; and Aila, T. 2019.
\newblock A style-based generator architecture for generative adversarial
  networks.
\newblock In \emph{Proceedings of the IEEE Conference on Computer Vision and
  Pattern Recognition}.

\bibitem[{Karras et~al.(2020)Karras, Laine, Aittala, Hellsten, Lehtinen, and
  Aila}]{sg2}
Karras, T.; Laine, S.; Aittala, M.; Hellsten, J.; Lehtinen, J.; and Aila, T.
  2020.
\newblock Analyzing and improving the image quality of {StyleGAN}.
\newblock In \emph{Proceedings of the IEEE Conference on Computer Vision and
  Pattern Recognition}.

\bibitem[{Kynk\"{a}\"{a}nniemi et~al.(2019)Kynk\"{a}\"{a}nniemi, Karras, Laine,
  Lehtinen, and Aila}]{IPR}
Kynk\"{a}\"{a}nniemi, T.; Karras, T.; Laine, S.; Lehtinen, J.; and Aila, T.
  2019.
\newblock Improved precision and recall metric for assessing generative models.
\newblock In \emph{Advances in Neural Information Processing Systems},
  volume~32.

\bibitem[{Kynkäänniemi et~al.(2022)Kynkäänniemi, Karras, Aittala, Aila, and
  Lehtinen}]{role}
Kynkäänniemi, T.; Karras, T.; Aittala, M.; Aila, T.; and Lehtinen, J. 2022.
\newblock The role of {ImageNet} classes in {Fréchet} {Inception} distance.
\newblock arXiv:2203.06026.

\bibitem[{Lee and Lee(2021)}]{trend}
Lee, J.; and Lee, J.-S. 2021.
\newblock {TREND}: Truncated generalized normal density estimation of
  {I}nception embeddings for {GAN} evaluation.
\newblock arXiv:2104.14767.

\bibitem[{Liu et~al.(2021)Liu, Lin, Cao, Hu, Wei, Zhang, Lin, and Guo}]{swin}
Liu, Z.; Lin, Y.; Cao, Y.; Hu, H.; Wei, Y.; Zhang, Z.; Lin, S.; and Guo, B.
  2021.
\newblock Swin {Transformer}: Hierarchical vision {Transformer} using shifted
  windows.
\newblock In \emph{Proceedings of the International Conference on Computer
  Vision}, 10012--10022.

\bibitem[{Morozov, Voynov, and Babenko(2021)}]{fidself}
Morozov, S.; Voynov, A.; and Babenko, A. 2021.
\newblock On self-supervised image representations for {GAN} evaluation.
\newblock In \emph{Proceedings of the International Conference on Learning
  Representations}.

\bibitem[{Naeem et~al.(2020)Naeem, Oh, Uh, Choi, and Yoo}]{DC}
Naeem, M.~F.; Oh, S.~J.; Uh, Y.; Choi, Y.; and Yoo, J. 2020.
\newblock Reliable fidelity and diversity metrics for generative models.
\newblock In \emph{Advances in Neural Information Processing Systems}, volume
  119, 7176--7185.

\bibitem[{Parmar, Zhang, and Zhu(2022)}]{aliased}
Parmar, G.; Zhang, R.; and Zhu, J.-Y. 2022.
\newblock On aliased resizing and surprising subtleties in {GAN} evaluation.
\newblock In \emph{Proceedings of the IEEE Conference on Computer Vision and
  Pattern Recognition}, 11410--11420.

\bibitem[{Paszke et~al.(2019)Paszke, Gross, Massa, Lerer, Bradbury, Chanan,
  Killeen, Lin, Gimelshein, Antiga, Desmaison, Kopf, Yang, DeVito, Raison,
  Tejani, Chilamkurthy, Steiner, Fang, Bai, and Chintala}]{pytorch}
Paszke, A.; Gross, S.; Massa, F.; Lerer, A.; Bradbury, J.; Chanan, G.; Killeen,
  T.; Lin, Z.; Gimelshein, N.; Antiga, L.; Desmaison, A.; Kopf, A.; Yang, E.;
  DeVito, Z.; Raison, M.; Tejani, A.; Chilamkurthy, S.; Steiner, B.; Fang, L.;
  Bai, J.; and Chintala, S. 2019.
\newblock PyTorch: An imperative style, high-performance deep learning library.
\newblock In \emph{Advances in Neural Information Processing Systems},
  volume~32.

\bibitem[{Radford et~al.(2021)Radford, Kim, Hallacy, Ramesh, Goh, Agarwal,
  Sastry, Askell, Mishkin, Clark et~al.}]{clip}
Radford, A.; Kim, J.~W.; Hallacy, C.; Ramesh, A.; Goh, G.; Agarwal, S.; Sastry,
  G.; Askell, A.; Mishkin, P.; Clark, J.; et~al. 2021.
\newblock Learning transferable visual models from natural language
  supervision.
\newblock In \emph{Proceedings of the International Conference on Machine
  Learning}, 8748--8763.

\bibitem[{Ramanujan et~al.(2020)Ramanujan, Wortsman, Kembhavi, Farhadi, and
  Rastegari}]{Ramanujan_2020_CVPR}
Ramanujan, V.; Wortsman, M.; Kembhavi, A.; Farhadi, A.; and Rastegari, M. 2020.
\newblock What's hidden in a randomly weighted neural network?
\newblock In \emph{Proceedings of the IEEE Conference on Computer Vision and
  Pattern Recognition}.

\bibitem[{Sajjadi et~al.(2018)Sajjadi, Bachem, Lucic, Bousquet, and Gelly}]{PR}
Sajjadi, M. S.~M.; Bachem, O.; Lucic, M.; Bousquet, O.; and Gelly, S. 2018.
\newblock Assessing generative models via precision and recall.
\newblock In \emph{Advances in Neural Information Processing Systems},
  5234--5243.

\bibitem[{Salimans et~al.(2016)Salimans, Goodfellow, Zaremba, Cheung, Radford,
  Chen, and Chen}]{IS}
Salimans, T.; Goodfellow, I.; Zaremba, W.; Cheung, V.; Radford, A.; Chen, X.;
  and Chen, X. 2016.
\newblock Improved techniques for training {GANs}.
\newblock In \emph{Advances in Neural Information Processing Systems},
  2234--2242.

\bibitem[{Sauer et~al.(2021)Sauer, Chitta, M{\"{u}}ller, and Geiger}]{projgan}
Sauer, A.; Chitta, K.; M{\"{u}}ller, J.; and Geiger, A. 2021.
\newblock Projected {GANs} converge faster.
\newblock In \emph{Advances in Neural Information Processing Systems}.

\bibitem[{Sauer, Schwarz, and Geiger(2022)}]{sgxl}
Sauer, A.; Schwarz, K.; and Geiger, A. 2022.
\newblock {StyleGAN-XL}: Scaling {StyleGAN} to large diverse datasets.
\newblock In \emph{Special Interest Group on Computer Graphics and Interactive
  Techniques Conference Proceedings}, 1--10.

\bibitem[{Seitzer(2020)}]{pytorch_fid}
Seitzer, M. 2020.
\newblock {pytorch-fid: FID score for PyTorch}.
\newblock \url{https://github.com/mseitzer/pytorch-fid}.

\bibitem[{Simonyan and Zisserman(2015)}]{vgg}
Simonyan, K.; and Zisserman, A. 2015.
\newblock Very deep convolutional networks for large-scale image recognition.
\newblock In \emph{Proceedings of the International Conference on Learning
  Representations}.

\bibitem[{Szegedy et~al.(2016)Szegedy, Vanhoucke, Ioffe, Shlens, and
  Wojna}]{Inception}
Szegedy, C.; Vanhoucke, V.; Ioffe, S.; Shlens, J.; and Wojna, Z. 2016.
\newblock Rethinking the Inception architecture for computer vision.
\newblock In \emph{Proceedings of the IEEE Conference on Computer Vision and
  Pattern Recognition}, 2818--2826.

\bibitem[{Theis, van~den Oord, and Bethge(2016)}]{theis2016note}
Theis, L.; van~den Oord, A.; and Bethge, M. 2016.
\newblock A note on the evaluation of generative models.
\newblock In \emph{Proceedings of the International Conference on Learning
  Representations}.

\bibitem[{Ulyanov, Vedaldi, and Lempitsky(2018)}]{dip}
Ulyanov, D.; Vedaldi, A.; and Lempitsky, V. 2018.
\newblock Deep image prior.
\newblock In \emph{Proceedings of the IEEE Conference on Computer Vision and
  Pattern Recognition}, 9446--9454.

\bibitem[{Wightman(2019)}]{timm}
Wightman, R. 2019.
\newblock PyTorch image models.
\newblock \url{https://github.com/rwightman/pytorch-image-models}.

\end{thebibliography}


\appendix
\setcounter{figure}{0}
\setcounter{table}{0}
\renewcommand{\thefigure}{\Alph{figure}}
\renewcommand{\thetable}{\Alph{table}}

\hyphenation{Microsoft}

\section{Implementation Details}
The experiments are run on a PC with a 3.7GHz Intel Core i7-8700K CPU, a 8GB NVIDIA RTX2070 Super GPU, and 32GB RAM.
We use the \texttt{PyTorch} 1.10.1~\cite{pytorch} on the Microsoft Windows~10 operating system.
We use the \texttt{timm} and \texttt{pytorch-fid} packages for the models with random weights and the trained Inception model, respectively~\cite{timm,pytorch_fid}.
We use the official implementation of the SwAV model with the ResNet50 structure~\cite{swav}.
To set a random seed for weight initialization, we use the \verb+torch.manual_seed(s)+ function with \texttt{s} $\in\{0,1,2,3,4\}$.

\section{Further Results of Nearest Neighbor Retrieval}

\begin{figure}[t]
     \centering
     \begin{subfigure}[b]{\linewidth}
         \centering
         \includegraphics[width=0.93\textwidth]{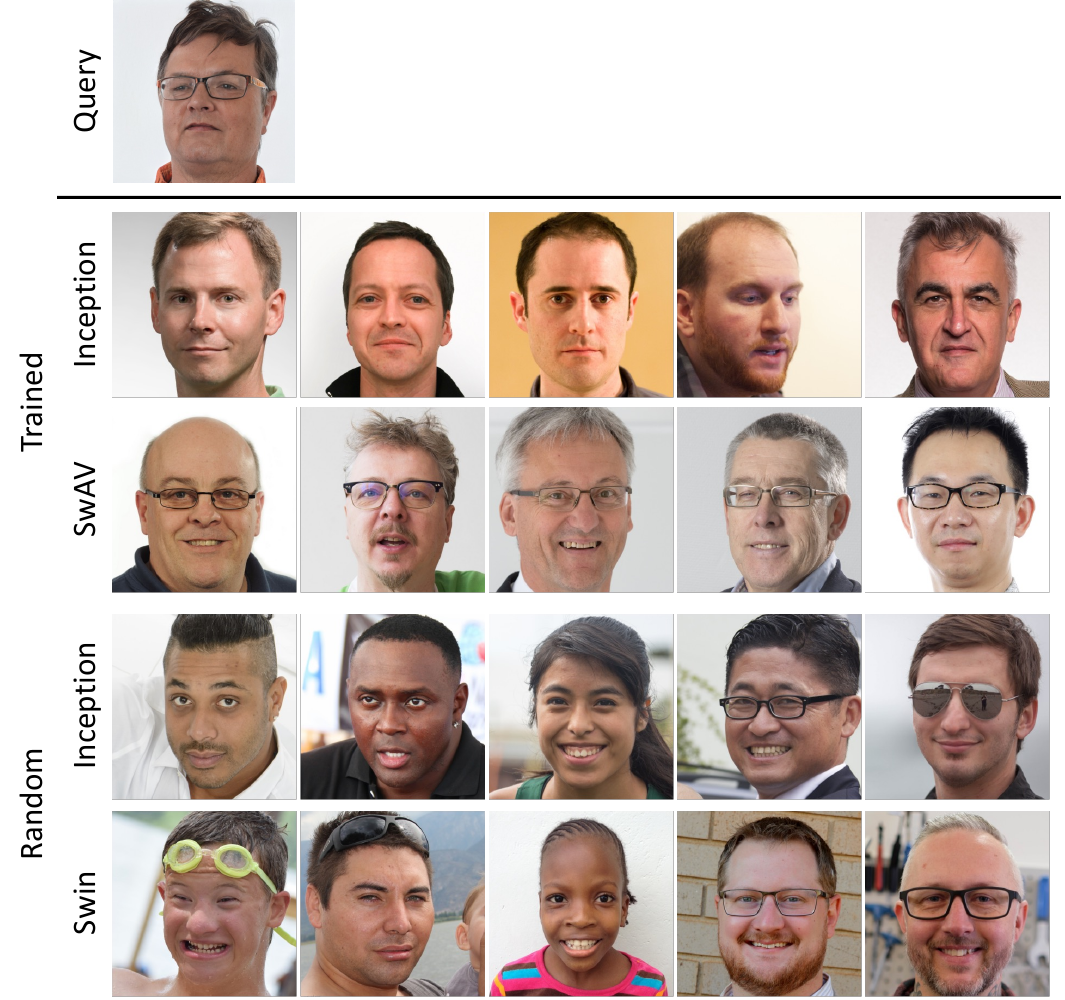}
         \caption{}
         \label{fig:nn2}
     \end{subfigure}
     \begin{subfigure}[b]{\linewidth}
         \centering
         \includegraphics[width=0.93\textwidth]{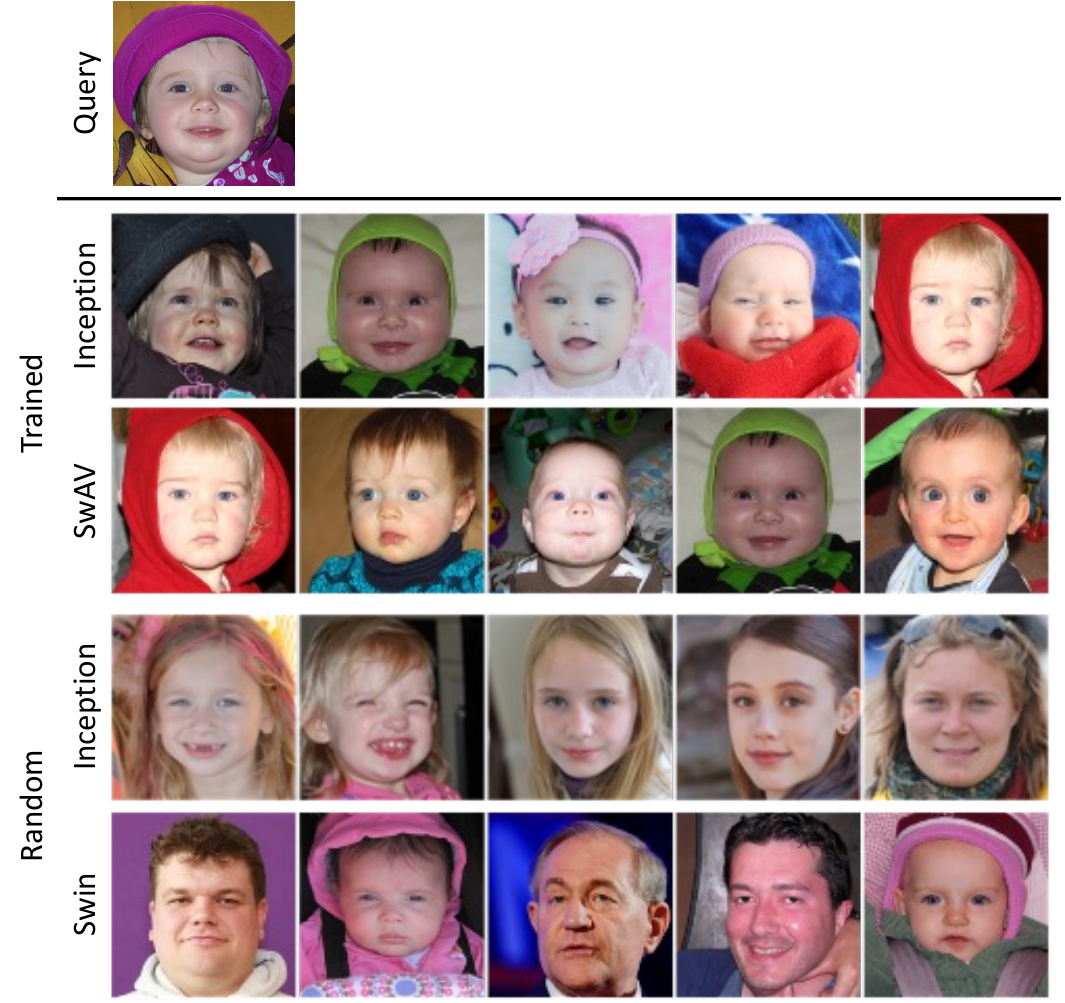}
         \caption{}
         \label{fig:nn3}
     \end{subfigure}
        \caption{Five nearest neighbors among the ground truth real images by each model for two generated images as query}
        \label{fig:nn-supp}
\end{figure}

To supplement the results of nearest neighbor retrieval, we provide additional results with different query images in \figurename~\ref{fig:nn-supp}.
In the case of the trained Inception, the queries and retrieved images share some high-level semantics such as adult males (\figurename~\ref{fig:nn2}) and babies wearing bonnets (\figurename~\ref{fig:nn3}).
It is notable that class labels of the ImageNet dataset include bonnets but not glasses, thus the retrieved images for the trained Inception in \figurename~\ref{fig:nn2} do not contain any glasses.
SwAV also retrieves images with similar semantics i.e., adult males wearing glasses and babies in \figurename s~\ref{fig:nn2} and \ref{fig:nn3}, respectively.
In the cases of both random features, the retrieved images sometimes include glasses and babies, but the features mostly focus on low-level information such as brightness and color.

\section{Variation of FID across Five Runs}

\begin{table*}[t]
\centering
\scriptsize
\begin{tabular}{@{}lrrcccc@{}}
\toprule
\textbf{Datasets} & \multicolumn{2}{c}{Trained} & \multicolumn{4}{c}{Random} \\ \cmidrule(l){2-3} \cmidrule(l){4-7}
Models & Inception & SwAV & Inception & VGG & Swin & ViT \\ \midrule
\textbf{ImageNet $256^2$}&        &      & \multicolumn{2}{l}{\hspace{12.5mm}$\times$10\textsuperscript{-15}}\\
BigGAN            & 6.24      & 4.45     &  2.73    $\pm$ 1.80   & 0.009 $\pm$ 0.003  &  1.07 $\pm$ 0.02 & 0.62 $\pm$ 0.04   \\
ADM-U             & 4.94      & 2.25     &  8.52    $\pm$ 4.48   & 0.012 $\pm$ 0.004  &  3.57 $\pm$ 0.14 & 1.61 $\pm$ 0.09   \\ \midrule
\textbf{ImageNet $64^2$}&         &      &              &    &              \\
E-VDVAE           & 46.23     & 6.52     &  367.68  $\pm$ 170.54 & 0.279 $\pm$ 0.086  &  4.58 $\pm$ 0.27 & 2.17 $\pm$ 0.03     \\
ADM-C             & 8.91      & 1.24     &  28.95   $\pm$ 15.26  & 0.021 $\pm$ 0.006  &  2.92 $\pm$ 0.12 & 1.23 $\pm$ 0.07     \\ \midrule
\textbf{FFHQ $1024^2$}&           &      &              &    &              \\
StyleGAN          & 4.58      & 0.92     &  4.62    $\pm$ 3.76   & 0.008 $\pm$ 0.003  &  0.34 $\pm$ 0.03 & 0.18 $\pm$ 0.02     \\
StyleGAN2         & 3.10      & 0.42     &  0.49    $\pm$ 0.27   & 0.005 $\pm$ 0.002  &  0.23 $\pm$ 0.01 & 0.13 $\pm$ 0.02     \\ \midrule
\textbf{FFHQ $256^2$} &           &      &              &    &              \\
E-VDVAE           & 72.10     & 3.42     &  321.11  $\pm$ 152.86 & 0.447 $\pm$ 0.145  &  7.87 $\pm$ 0.12 & 4.87 $\pm$ 0.77     \\
ProjectedGAN      & 4.18      & 1.32     &  18.01   $\pm$ 11.72  & 0.028 $\pm$ 0.008  &  0.68 $\pm$ 0.02 & 0.35 $\pm$ 0.04     \\ \midrule
\textbf{\pokemon $256^2$}  &           &      &         &         &              \\
ProjectedGAN      & 27.88     & 4.80     &  90.40    $\pm$ 47.42  & 0.343 $\pm$ 0.086  &  3.86 $\pm$ 0.11 & 2.79 $\pm$ 0.18     \\
StyleGAN-XL       & 25.46     & 3.43     &  31.22    $\pm$ 19.35  & 0.271 $\pm$ 0.073  &  1.12 $\pm$ 0.03 & 0.67 $\pm$ 0.07     \\
\bottomrule
\end{tabular}
\caption{Evaluation results in terms of FID using different feature spaces for various generative models. For the models with random weights, the average and standard deviation of the scores are shown.}
\label{tab:eval-FID-var} 
\end{table*}

\tablename~\ref{tab:eval-FID-var} shows the evaluation results of generative models using different features for FID.
For the random features, we show the average and standard deviation of FID across five runs using different random seeds.
As shown in \figurename s~6a and 6b, the standard deviation values of random CNN features are quite large, reaching nearly a half of the average values.
On the other hand, random features from Transformer structures yield much smaller standard deviation values.
While some standard deviation values are quite large, we observe that the superiority between the generative models do not change in each run for all random features.

\section{Other Metrics}

\subsection{Kernel Inception Distance}

\begin{table*}[t]
\centering
\scriptsize
\begin{tabular}{@{}lrrrrrr@{}}
\toprule
\textbf{Datasets} & \multicolumn{2}{c}{Trained} & \multicolumn{4}{c}{Random} \\ \cmidrule(l){2-3} \cmidrule(l){4-7}
Models & Inception & SwAV & Inception & VGG & Swin & ViT \\ \midrule
\textbf{ImageNet $256^2$}&        &      & $\times$10\textsuperscript{-14} & $\times$10\textsuperscript{-6} \\
BigGAN            & 0.0009      & 0.0012     &  2.23       & 0.54   &  0.004 & 0.015    \\
ADM-U             & 0.0016      & 0.0005     &  2.22       & 1.67   &  0.013 & 0.033    \\ \midrule
\textbf{ImageNet $64^2$}&\\
E-VDVAE           & 0.0403      & 0.0026     &  1.77       & 190.78  & 0.022  & 0.048       \\
ADM-C             & 0.0081      & 0.0005     &  1.89       & 5.11    & 0.009  & 0.024       \\ \midrule
\textbf{FFHQ $1024^2$}&\\
StyleGAN          & 0.0011      & 0.0004     &  2.08       & 1.06   &  0.001   & 0.003      \\
StyleGAN2         & 0.0006      & $<$0.0001  &  2.09       & 0.32   &  $<$0.001& 0.003      \\ \midrule
\textbf{FFHQ $256^2$} &\\
E-VDVAE           & 0.0620      & 0.0020     &  2.01       & 285.52 &  0.031 & 0.128      \\
ProjectedGAN      & 0.0007      & 0.0008     &  2.10       & 3.03   &  0.002 & 0.006      \\ \midrule
\textbf{\pokemon $256^2$}  &\\
ProjectedGAN      & 0.0035     & 0.0035     &  0.41        & 33.43   &  0.007 & 0.016      \\
StyleGAN-XL       & 0.0016     & 0.0021     &  0.40        & 11.06   &  0.001 & 0.003     \\
\bottomrule
\end{tabular}
\caption{Evaluation results in terms of KID using different feature spaces for various generative models}
\label{tab:eval-KID} 
\end{table*}

\citeauthor{KID}~\shortcite{KID} proposed the Kernel Inception distance (KID) that measures the maximum mean discrepancy between real and generated images after transforming Inception features using a kernel function.
\tablename~\ref{tab:eval-KID} shows the evaluation results of generative models using KID with the trained and random features.
We use a polynomial kernel of order 3, which is most widely used.
In most cases, the results of KID are consistent with those of FID in \tablename~2.

\subsection{Precision and Recall}

\begin{table*}[t]
\centering
\scriptsize
\begin{tabular}{@{}lcccccc@{}}
\toprule
\textbf{Datasets} & \multicolumn{2}{c}{Trained} & \multicolumn{4}{c}{Random} \\ \cmidrule(l){2-3} \cmidrule(l){4-7}
Models & Inception & SwAV & Inception & VGG & Swin & ViT \\ \midrule
\textbf{\pokemon $256^2$}  &           &      &         &         &              \\
ProjectedGAN      & 0.813 / 0.433      & 0.809 / 0.012      & 0.873 / 0.847   & 0.474 / 0.448   &  0.759 / 0.249 & 0.720 / 0.419      \\
StyleGAN-XL       & 0.813 / 0.515      & 0.906 / 0.066      & 0.915 / 0.881   & 0.577 / 0.628   &  0.791 / 0.190 & 0.775 / 0.520      \\
\bottomrule
\end{tabular}
\caption{Evaluation results in terms of Precision / Recall}
\label{tab:eval-PR} 
\end{table*}

We use the improved precision and recall metrics~\cite{IPR} using random features.
\tablename~\ref{tab:eval-PR} shows the evaluation results for ProjectedGAN and StyleGAN-XL trained on the \pokemon dataset.
In most cases, the results of both Precision and Recall are in line with those of FID.
The superiority between the generative models are consistent over all features except for Recall using random Swin.
In the case of SwAV, Recall for both GANs are quite low, which we think that the GANs are underrated although they actually generate diverse images as shown in~\cite{projgan,sgxl}.

\section{BigGAN Outliers}
\begin{figure}[t]
     \centering
     \begin{subfigure}[b]{\linewidth}
         \centering
         \includegraphics[width=0.9\textwidth]{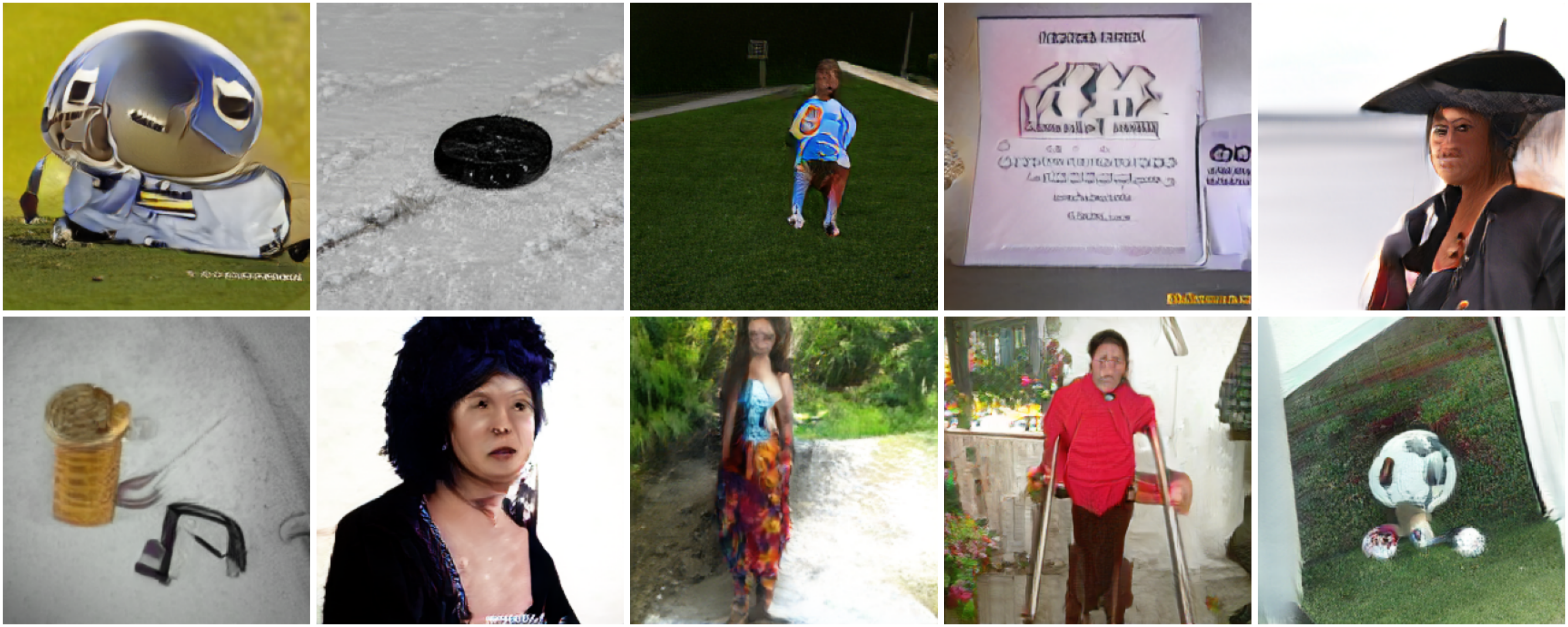}
         \caption{High-level outliers}
         \label{fig:bg-high}
     \end{subfigure}
     \begin{subfigure}[b]{\linewidth}
         \centering
         \includegraphics[width=0.9\textwidth]{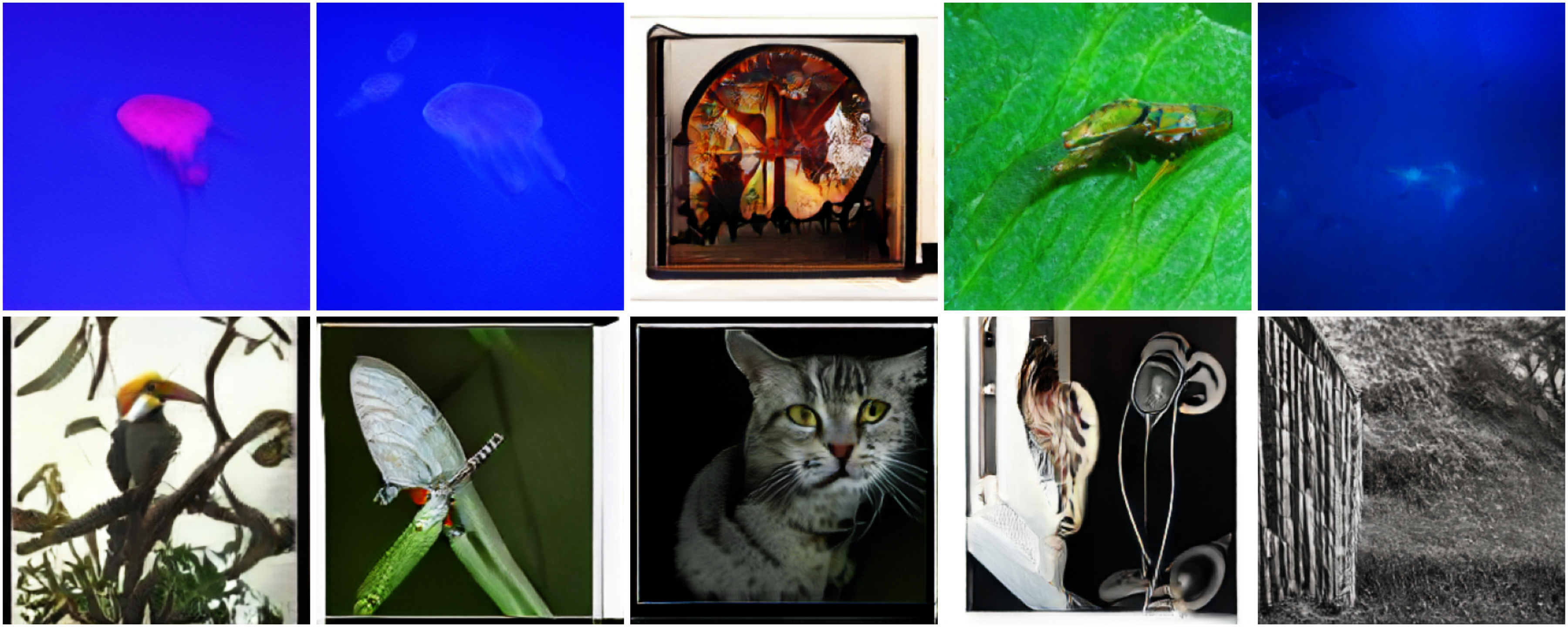}
         \caption{Low-level outliers}
         \label{fig:bg-low}
     \end{subfigure}
        \caption{BigGAN outliers using $d_\mathtt{CLIP}$ and $d_\mathtt{STEM}$}
        \label{fig:bg-outliers}
\end{figure}

\figurename~\ref{fig:bg-outliers} shows the high-level and low-level outliers generated by BigGAN trained on ImageNet using $d_\mathtt{CLIP}$ and $d_\mathtt{STEM}$.
The overall trend is similar to that in \figurename~3.
For the high-level outliers (\figurename~\ref{fig:bg-high}), we can observe severe errors in high-level semantics such as unidentifiable objects.
In \figurename~\ref{fig:bg-low}, the low-level outliers contain unnatural colors, while they often contain objects with acceptable quality.

\begin{figure}[t]
     \centering
     \begin{subfigure}[b]{0.4\linewidth}
         \centering
         \includegraphics[width=\textwidth]{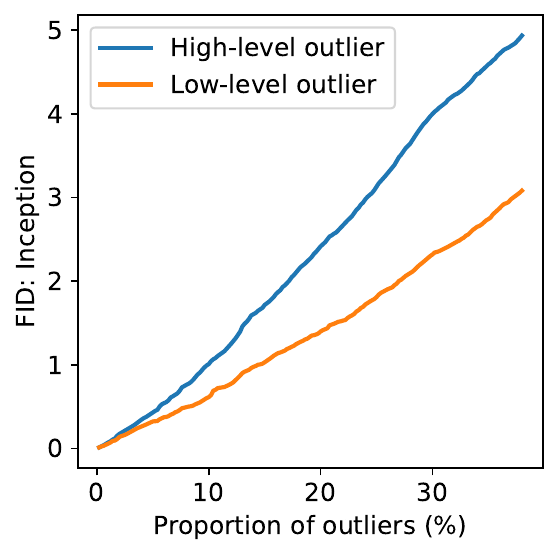}
         \caption{Inception}
         \label{fig:replace-bg-inception}
     \end{subfigure}
     \hspace{2.4mm}
     \begin{subfigure}[b]{0.42\linewidth}
         \centering
         \includegraphics[width=\textwidth]{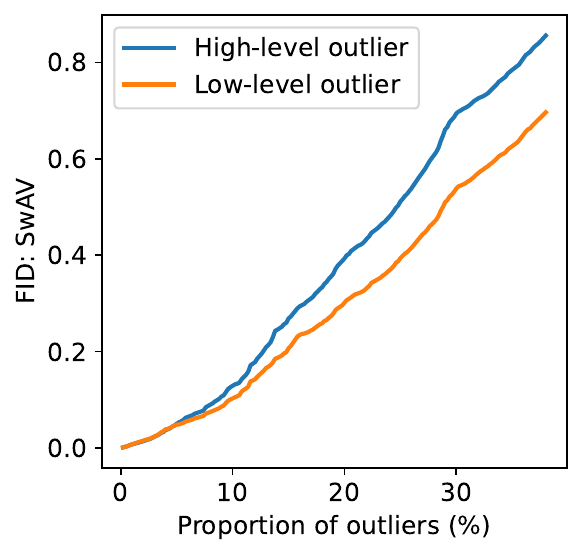}
         \caption{SwAV}
         \label{fig:replace-bg-swav}
     \end{subfigure}
     \begin{subfigure}[b]{0.43\linewidth}
         \centering
         \includegraphics[width=\textwidth]{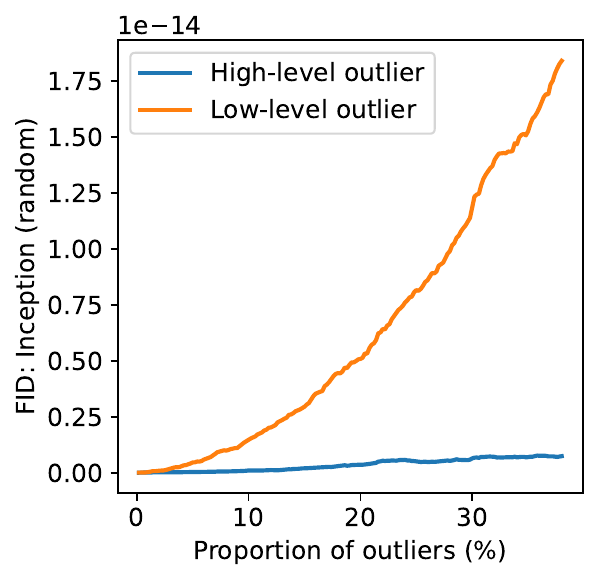}
         \caption{Inception (random)}
         \label{fig:replace-bg-inception-r}
     \end{subfigure}
     \hspace{0.6mm}
     \begin{subfigure}[b]{0.45\linewidth}
         \centering
         \includegraphics[width=\textwidth]{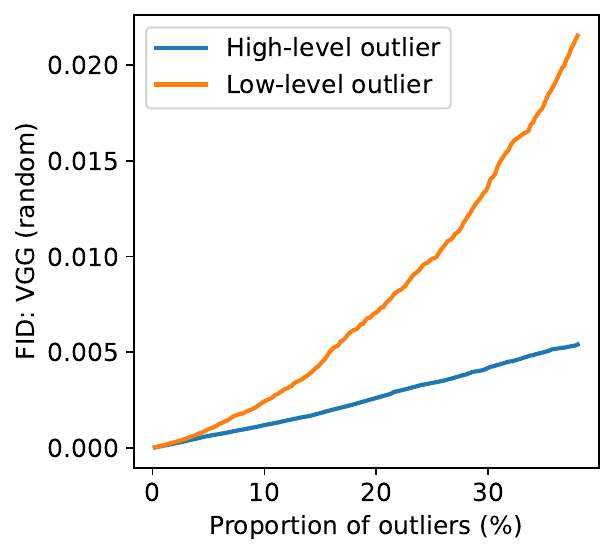}
         \caption{VGG (random)}
         \label{fig:replace-bg-vgg-r}
     \end{subfigure}
     \begin{subfigure}[b]{0.42\linewidth}
         \centering
         \includegraphics[width=\textwidth]{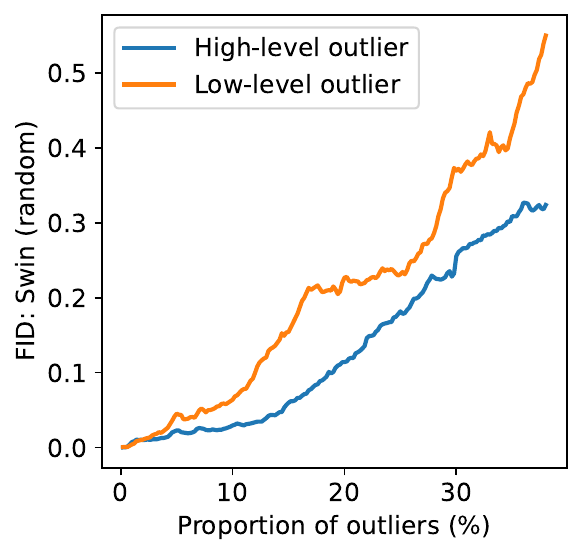}
         \caption{Swin (random)}
         \label{fig:replace-bg-swin-r}
     \end{subfigure}
     \hspace{2.2mm}
     \begin{subfigure}[b]{0.43\linewidth}
         \centering
         \includegraphics[width=\textwidth]{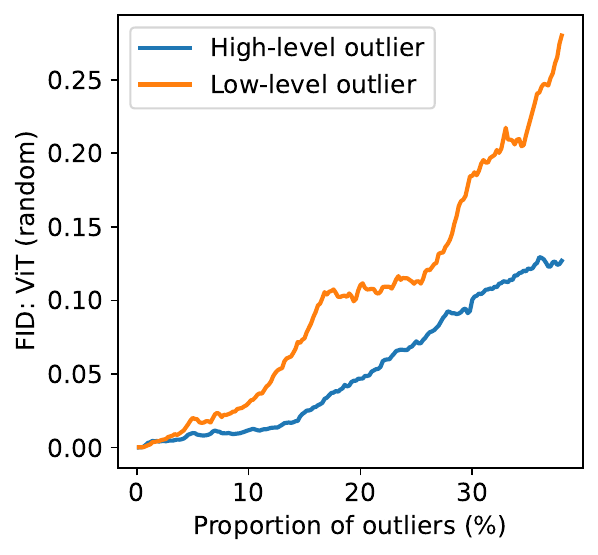}
         \caption{ViT (random)}
         \label{fig:replace-bg-vit-r}
     \end{subfigure}
     \hspace{0.4mm}
        \caption{FID vs. the proportion of outliers of BigGAN for different features}
        \label{fig:replace-bg}
\end{figure}

Similarly to \figurename~4, \figurename~\ref{fig:replace-bg} shows the FID with respect to the proportion of the outliers, when the ImageNet images are gradually replaced with the high-level or low-level outliers of BigGAN.
As in \figurename~4, the trained features are more sensitive to the high-level outliers and the random features are more sensitive to the low-level outliers.

\end{document}